\documentclass[10pt,twocolumn,letterpaper]{article}

\usepackage{cvpr}              

\usepackage[dvipsnames]{xcolor}

\definecolor{cvprblue}{rgb}{0.21,0.49,0.74}
\usepackage[pagebackref,breaklinks,colorlinks,allcolors=cvprblue]{hyperref}
\usepackage{booktabs}
\usepackage{multirow}
\usepackage{algorithm}
\usepackage{algorithmic}
\usepackage{graphicx, amsmath, amssymb, caption, subcaption, multirow, overpic, textpos}
\usepackage[dvipsnames]{xcolor}
\usepackage{arydshln}
\usepackage{colortbl}
\usepackage{textcomp}
\usepackage{pifont}
\usepackage{xcolor}

\definecolor{lightgray}{gray}{0.7}

\def\ie{\emph{i.e}\onedot} 
 
\def\etc{\emph{etc}\onedot} \def\vs{\emph{vs}\onedot}

\newlength\savewidth\newcommand\shline{\noalign{\global\savewidth\arrayrulewidth
  \global\arrayrulewidth 1pt}\hline\noalign{\global\arrayrulewidth\savewidth}}
\newcommand{\tablestyle}[2]{\setlength{\tabcolsep}{#1}\renewcommand{\arraystretch}{#2}\centering\footnotesize}

\definecolor{baselinecolor}{gray}{.9}

\definecolor{darkergreen}{RGB}{10,100,10}
\newcommand{\cmark}{\textcolor{darkergreen}{\ding{51}}}
\newcommand{\xmark}{\textcolor{red}{\ding{55}}}

\def\confName{CVPR}
\def\confYear{2025}

\title{COCONut-PanCap: Joint Panoptic Segmentation and Grounded Captions for Fine-Grained Understanding and Generation}

\author{
Xueqing Deng%
\quad
Qihang Yu%
\quad
Ali Athar%
\quad
Chenglin Yang%
\quad
Linjie Yang\\[5pt]
\quad
Xiaojie Jin%
\quad
Xiaohui Shen%
\quad
Liang-Chieh Chen\\[5pt]
ByteDance Seed\\[5pt]
\href{https://xdeng7.github.io/coconut.github.io/coconut_pancap.html}{Project Page}\\[5pt]
{\tt\small xueqingdeng@bytedance.com}\\
}

\begin{document}
\twocolumn[{%
\maketitle\centering
\vspace{-20pt}
\includegraphics[width=\linewidth]{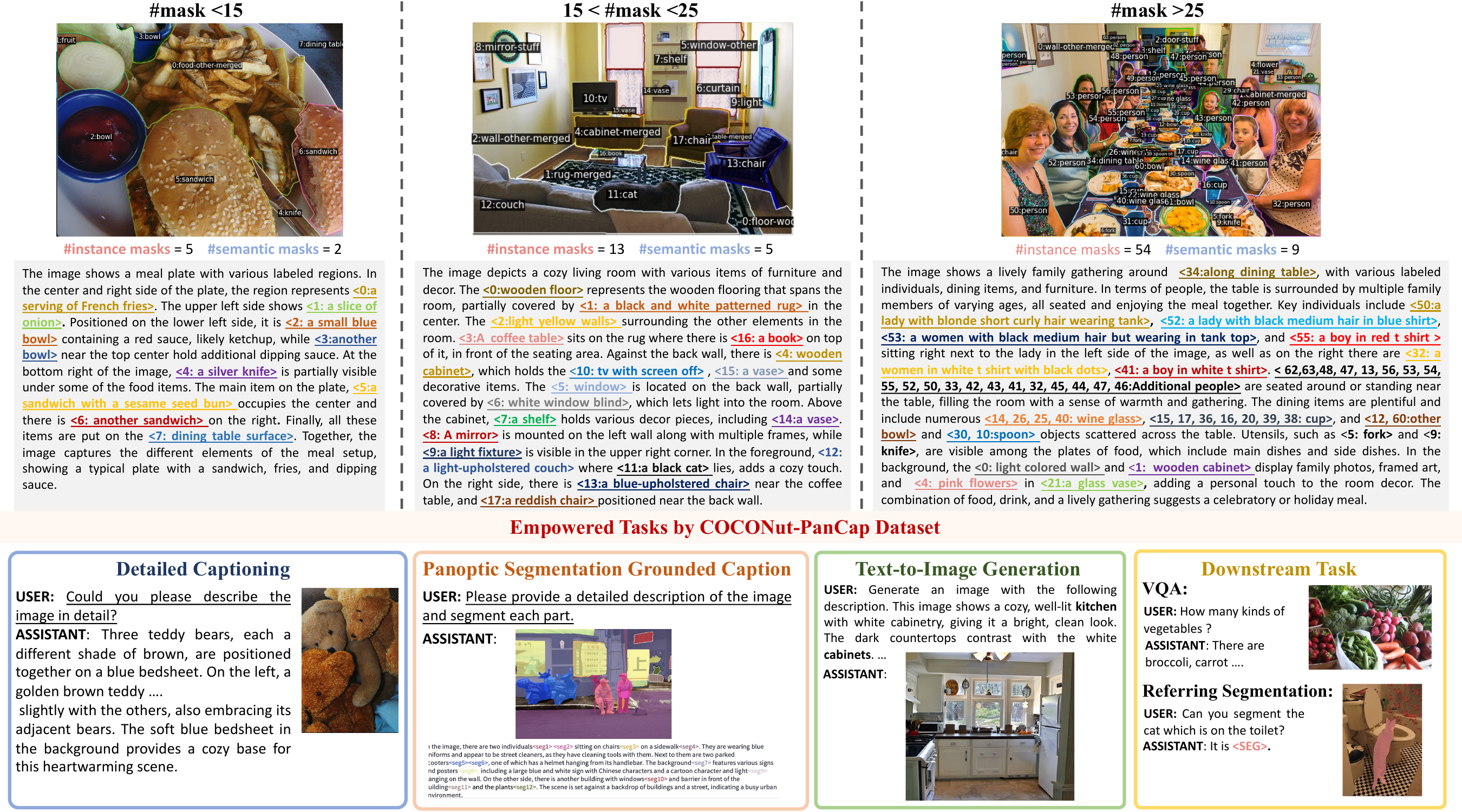}
    \vspace{-20pt}
  \captionof{figure}{
  \textbf{COCONut-PanCap Dataset.}
  \textit{Top:} The proposed COCONut-PanCap dataset features detailed captions grounded with dense panoptic segmentation masks.
  \textit{Bottom:} COCONut-PanCap supports various fine-grained understanding and generation tasks, including detailed captioning, panoptic segmentation grounded caption, and text-to-image generation.
  The dataset also facilitates several downstream tasks, such as visual question-answering (VQA) and referring segmentation.
  }
    \vspace{10pt}
    \label{fig:teaser}
}]

\begin{abstract}
This paper introduces the COCONut-PanCap dataset, created to enhance panoptic segmentation and grounded image captioning. Building upon the COCO dataset with advanced COCONut panoptic masks, this dataset aims to overcome limitations in existing image-text datasets that often lack detailed, scene-comprehensive descriptions. The COCONut-PanCap dataset incorporates fine-grained, region-level captions grounded in panoptic segmentation masks, ensuring consistency and improving the detail of generated captions.
Through human-edited, densely annotated descriptions, COCONut-PanCap supports improved training of vision-language models (VLMs) for image understanding and generative models for text-to-image tasks.
Experimental results demonstrate that COCONut-PanCap significantly boosts performance across understanding and generation tasks, offering complementary benefits to large-scale datasets. This dataset sets a new benchmark for evaluating models on joint panoptic segmentation and grounded captioning tasks, addressing the need for high-quality, detailed image-text annotations in multi-modal learning. 
\end{abstract}    
\section{Introduction}
Recent advancements in multi-modal foundation models have been largely driven by the availability of large-scale paired text-image datasets. These datasets, often collected via web crawling with basic filtering techniques~\cite{schuhmann2021laion400m,schuhmann2022laion5b,gadre2024datacomp}, contain low-quality, web-sourced captions that lack depth and accuracy. In contrast, human-annotated caption datasets, such as COCO-caption~\cite{chen2015coco_caption}, offer higher-quality descriptions but are limited in scale and tend to be concise, with an average caption length of 10 words. To overcome the limitations of short captions, the research community has leveraged vision-language models (VLMs)~\cite{liu2024llavanext,li2024recaption,li2024DenseFusion,chen2023sharegpt4v,team2023gemini} to generate detailed synthetic captions. While these machine-generated captions improve visual understanding~\cite{li2024DenseFusion,chen2023sharegpt4v} and generation tasks~\cite{li2024recaption}, they remain inferior to high-quality, human-verified annotations~\cite{Onoe2024docci}.

Addressing this challenge requires balancing scalability and annotation quality, as generating detailed and accurate image descriptions at scale remains labor-intensive~\cite{garg2024imageinwords,Onoe2024docci}. In this paper, we introduce an efficient annotation approach that combines dense mask annotations with commercial VLMs~\cite{chen2023sharegpt4v} to produce high-quality image captions. Our goal is to minimize human effort while generating rich, structured descriptions. 

To achieve this, we base our work on the COCO-caption dataset~\cite{chen2015coco_caption} due to its widespread use and diverse image content.
We revisit the COCO-caption dataset to provide more detailed and comprehensive caption annotations.
Our approach involves creating holistic captions synthesized from region-based dense captions that describe distinct areas within each image. Specifically, we build on recent COCONut panoptic segmentation annotations~\cite{deng2024coconut} to generate a new set of detailed captions by: (a) annotating each segmentation region with a VLM-generated draft, carefully refined through human corrections, and (b) summarizing these region captions into a comprehensive image caption while preserving the grounding correspondence between image masks and object references. This enables a novel task that integrates panoptic segmentation with grounded captioning.
Our structured annotation process ensures that the captions are both \textit{complete}, covering the majority of objects in each image, and \textit{grounded}, with precise segmentation masks.

The final dataset, named \textbf{COCONut-PanCap}, is designed for a wide range of vision-language applications, combining \textbf{Pan}optic segmentation and grounded \textbf{Cap}tioning. It comprises 118K image-text pairs for training, with an average caption length of 203 words, as well as an additional 25K image-text pairs, with an average caption length of 233 words for validation. We demonstrate that COCONut-PanCap significantly boosts the performance of both VLM and text-to-image generation models at the instruction tuning and fine-tuning stages, outperforming recent detailed caption datasets~\cite{Onoe2024docci}. This highlights the potential of our grounding-based captions for both vision-language understanding and image generation tasks.

Our contributions are summarized as follows: 

\begin{itemize}
    \item We propose a caption annotation pipeline leveraging panoptic segmentation to create a high-quality, detailed caption dataset comprising \textbf{143K} annotated images. The resulting annotations are comprehensive, accurate, and include grounding masks, making this dataset substantially larger than recent detailed caption datasets.
   
    \item Our \textbf{COCONut-PanCap} dataset facilitates a new challenging task combining \textbf{P}anoptic segmentation and \textbf{G}rounded \textbf{C}aptioning (\textbf{PGC}). We establish evaluation metrics and settings for this PGC task and benchmark several recent methods to assess performance on this novel challenge. 
    \item We validate the utility of our proposed dataset across various fine-grained Image-to-Text (I2T) and Text-to-Image (T2I) tasks, including detailed caption generation, PGC, visual question answering (VQA), referring segmentation, and text-conditioned image generation. Experimental results show that our dataset significantly enhances model performance across all these tasks.
\end{itemize}

\begin{table*}[t!]
\centering
\tablestyle{1.5pt}{1.05}
\begin{tabular}{l|r|c|r|c|c}
\textbf{Dataset Name} & \textbf{Image Source}&\textbf{Sample}  & \textbf{Annotated by}   & \textbf{Avg. Words} & \textbf{Masks} \\
\shline

BLIP-LCS & LAION~\cite{schuhmann2022laion5b}, CC~\cite{changpinyo2021cc12m}, SBU~\cite{vicente2011sbu} & 558K & BLIP~\cite{li2022blip} & 54 & \xmark \\
DenseFusion1M~\cite{li2024DenseFusion} & LAION~\cite{schuhmann2022laion5b} & 1,059K & Vision Specialist Models & 191 & \xmark \\
LLaVA-Recap118K~\cite{liu2024llavanext} & COCO~\cite{lin2014coco} & 118K & LLaVA-NEXT~\cite{liu2024llavanext} & 186 & \xmark \\
LLaVA-Details-23K~\cite{liu2023llava} & COCO~\cite{lin2014coco} & 23K & GPT4 & 105 & \xmark \\

ShareGPT4V~\cite{chen2023sharegpt4v} & LAION~\cite{schuhmann2022laion5b}, CC~\cite{changpinyo2021cc12m}, SBU~\cite{vicente2011sbu}, COCO~\cite{lin2014coco} \etc & 100K & GPT4-Vision & 162 & \xmark \\
ShareGPT4V-PT~\cite{chen2023sharegpt4v} & LAION~\cite{schuhmann2022laion5b}, CC~\cite{changpinyo2021cc12m}, SBU~\cite{vicente2011sbu}, COCO~\cite{lin2014coco} \etc & 1,246K & Share-Captioner~\cite{chen2023sharegpt4v} & 144 & \xmark \\
\hline
PixelLM-MUSE~\cite{ren2024pixellm} & LVIS~\cite{gupta2019lvis} & 246K & GPT4-Vision & - & 3.7\textsuperscript{\ddag}  \\
Osprey~\cite{yuan2024osprey} & COCO~\cite{lin2014coco} & 724K & GPT4-Vision & - & - \\
GLaMM-GCG~\cite{hanoona2023GLaMM} &  RefCOCOg~\cite{mao2016refcocog},PSG~\cite{yang2022psg},Flick30K~\cite{plummer2015flickr30k}  & 214K & Vision Specialist Models & 128 & 3.6 \\ 
\hline
COCO-caption~\cite{chen2015coco_caption} & COCO~\cite{lin2014coco} & 118K & \textcolor{darkergreen}{\textbf{Human}} & 11
& \xmark \\

DCI~\cite{Urbanek2024dci} & SA-1B~\cite{kirillov2023sam} & 8K & \textcolor{darkergreen}{\textbf{Human}}  & 144 & \xmark \\
DOCCI~\cite{Onoe2024docci} & DOCCI~\cite{Onoe2024docci} & 9.6K & \textcolor{darkergreen}{\textbf{Human}}  & 136 & \xmark \\
IIW~\cite{garg2024imageinwords} & WebLI~\cite{garg2024imageinwords} & 8.5K & \textcolor{darkergreen}{\textbf{Human}}  & 217 & \xmark \\
COCONut-PanCap (ours) & COCO~\cite{lin2014coco} & \textbf{118K} & \textcolor{darkergreen}{\textbf{Human}}  & 203 & \textbf{13.2} \\

\end{tabular}
\vspace{-10pt}
\caption{
\textbf{Dataset (training set) Comparison. }
Our proposed COCONut-PanCap dataset stands out for its \textbf{detailed} (2nd highest in Average Words), \textbf{high-quality} (human interactive annotated) \textit{captions} and \textbf{high-density} \textit{segmentation masks} (1st in Average Masks).
\textsuperscript{\ddag} denotes the mask number for referring segmentation which only counts the targets in QA format. Note that ``Samples" means the number of collected annotations, where there may exist one image with multiple different annotation, \ie, in region-level datasets like Osprey.
}
\label{tab:dataset_comp}
\end{table*}

\begin{table}[t!]
\centering
\tablestyle{1.0pt}{1.1}
\scalebox{0.93}{
\begin{tabular}{l|r|c|ccc}
\textbf{Dataset Name}       & \textbf{Samples} & \textbf{Avg. Words} & \textbf{Caption} & \textbf{T2I} & \textbf{Grd. Seg.} \\
\shline
COCO-30K~\cite{chen2015coco_caption}           & 30,000    & 11

& \cmark          & \cmark               &    \xmark                   \\
DOCCI-test~\cite{Onoe2024docci}         & 5,000     & 136

& \cmark          & \cmark                &    \xmark                    \\
IIW-test~\cite{garg2024imageinwords}           & 445       & 217

& \cmark            & \cmark                &    \xmark                    \\
GenEval~\cite{ghosh2023geneval}            & 553     & 8

&    \xmark         & \cmark                 &     \xmark                   \\
T2I-CompBench val ~\cite{huang2023t2icompbench}     & 2400     & 9

&    \xmark         & \cmark               &       \xmark                 \\
GLaMM-GCG val-test~\cite{hanoona2023GLaMM} & 2,000     & 128        & \cmark           &       \xmark            & \cmark                       \\
COCONut-PanCap val (ours) & 25,000    & 233         & \cmark            & \cmark                 & \cmark                     
\end{tabular}%
}
\caption{
\textbf{Dataset (evaluation set) Comparison.}
Our COCONut-PanCap validation set provides detailed captions and supports multiple multi-modal tasks, including image captioning, text-to-image generation (T2I), and grounded segmentation (Grd. Seg.).
}
\label{tab:eval_dataset_comp}
\end{table}

\section{Related Work}
\label{sec:related_work}

\noindent\textbf{Detailed Captions from VLMs.} Researchers are increasingly interested in creating large-scale datasets with detailed captions generated from advanced vision-language models. DenseFusion1M~\cite{li2024DenseFusion} utilizes a pretrained perceptual model to prompt VLMs, facilitating more detailed image descriptions. 

Recap-DataComp1B~\cite{li2024recaption}  first fine-tunes the Llama-3-8B powered LLaVA-1.5 model~\cite{liu2023improvedllava}, then applies it to recaption approximately 1.3 billion images from the DataComp-1B dataset~\cite{gadre2024datacomp}, generating a rich repository of detailed image descriptions. On a similar front, the PixelProse dataset~\cite{singla2024pixels} offers general-purpose image captions designed to serve various applications, from visual question answering (VQA) to pre-training tasks. Unlike datasets targeting single applications, PixelProse captions are dense, versatile image descriptions that can be adapted to other formats, such as VQA and instructional data, with the help of large language models (LLMs). Although these detailed caption datasets are large-scale, they are directly generated by VLMs without human verification, falling behind human-annotated captions on quality. Our proposed COCONut-PanCap dataset leverages extensive human effort to ensure high-quality annotations.

\noindent\textbf{Human-annotated Detailed Captions.} Several efforts have been made toward this goal, utilizing fully human-annotated data or human-in-the-loop approaches. One example is DOCCI ~\cite{Onoe2024docci} which is a small, high-detailed image caption dataset that is entirely human-annotated, containing only 15K samples but providing diverse details, such as key objects, their attributes, spatial relationships, and text rendering. Two small-scale detailed caption datasets, ImageInWords~\cite{garg2024imageinwords} and DCI~\cite{Urbanek2024dci}, use a combination of automatic annotation models with human involvement, both with fewer than 10K samples. Pixmo-Cap~\cite{deitke2024molmo} introduces a large-scale dataset of detailed image captions from speech-based descriptions, offering richer visual annotations than text-based methods. 

Our proposed COCONut-PanCap dataset yields smaller scale compare to Pixmo-Cap but we have different focuses where Pixmo-Cap focuses on pretraining the VLMs while we focus on the instruction tuning and finetuning stages of VLMs and image generation models. Our work also shares a similar annotation pipeline with a recent video captioning dataset Shot2Story~\cite{han2023shot2story20k} where both VLM draft and human corrections are used to create complete and accurate annotations.

\noindent\textbf{Grounded Captions with Segmentation Masks.}
Existing work have made significant strides in creating datasets with region-level captions linked to entity segmentation masks~\cite{yuan2024osprey} or bounding boxes~\cite{zhang2023gpt4roi}. However, few datasets associate grounded segmentation directly with captions. GLaMM~\cite{hanoona2023GLaMM} proposes a Grounding-anything Dataset (GranD) using an automated annotation pipeline that encompasses 7.5M unique concepts grounded in a total of 810M regions available with segmentation masks. 
 
Later, MGLMM~\cite{zhou2024mglmm} further explore the multi-granularity GLaMM model to generate a multi-granularity dataset. Our proposed COCONut-PanCap dataset follows a similar approach of grounding captions to dense masks but offers significantly denser masks per caption, as shown in Tab.~\ref{tab:dataset_comp}, with an average of 13.2 masks per image compared to 3.6 in GLaMM. Note that we focus on grounded segmentation for detailed captions, rather than descriptions of all levels of segmentation masks (objects or parts) as provided in the GranD dataset~\cite{hanoona2023GLaMM}, which is outside the scope of our study.

\begin{figure*}[ht!]
    \centering
    \includegraphics[width=\linewidth]{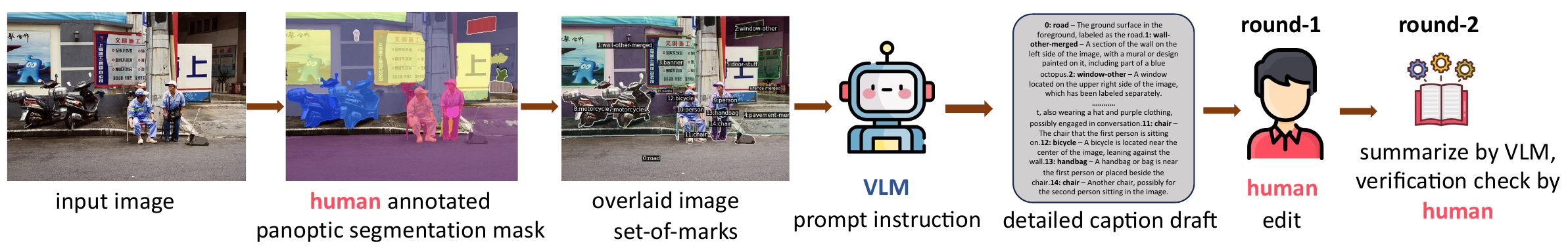}
    \caption{\textbf{Annotation Pipeline.} Given an input image, human-annotated panoptic segmentation masks are overlaid using set-of-marks~\cite{yang2023setofmark} visualization techniques to prompt the vision-language model (VLM). After generating an initial draft, human effort is investigated for editing and verification. Finally, the annotated metadata will be formatted to construct the datasets for various tasks at instruction tuning or finetuning stage.}
    \label{fig:pipeline}
\end{figure*}

\begin{figure*}[ht!]
    \centering
    \includegraphics[width=\linewidth]{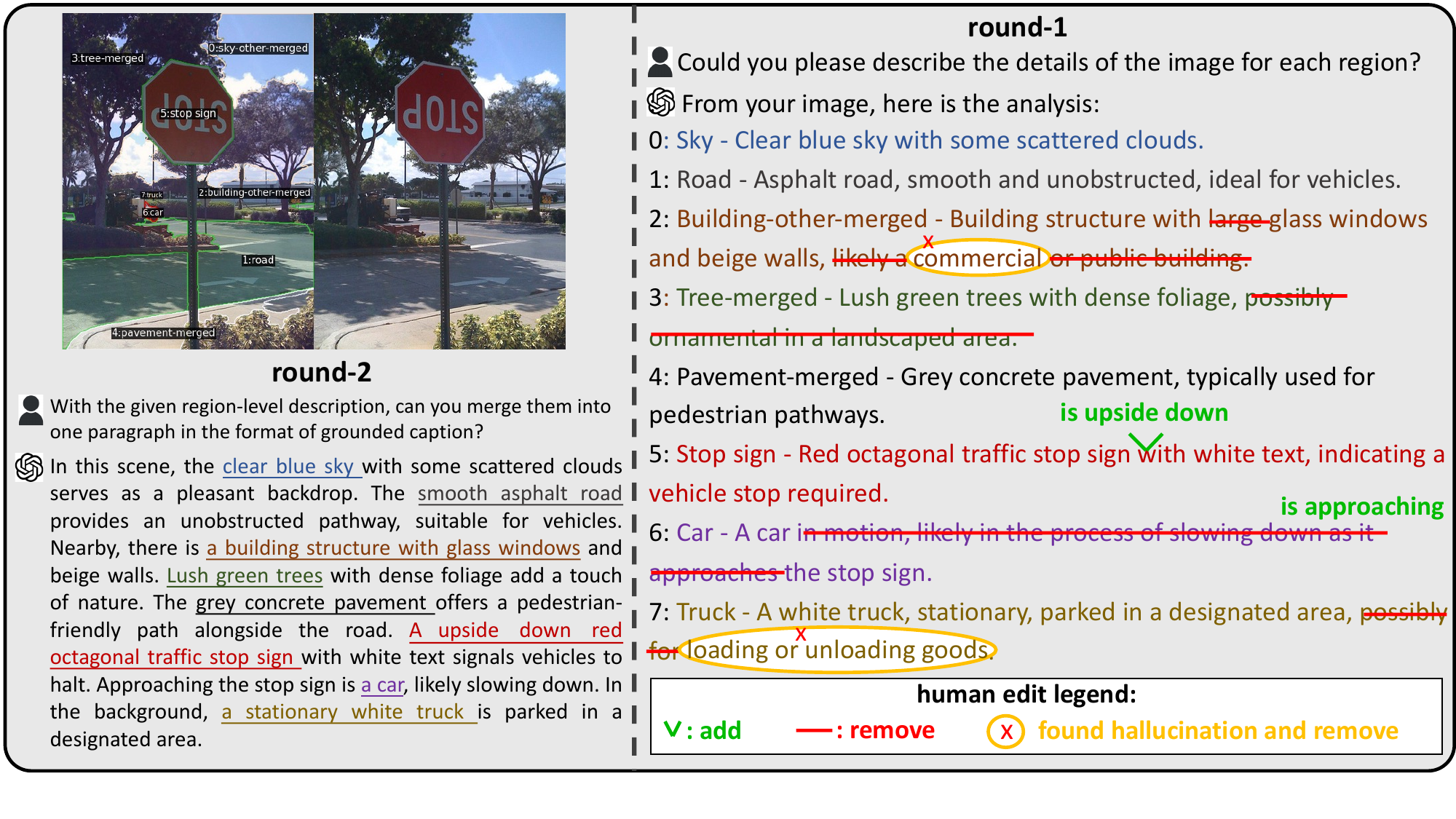}
    \caption{\textbf{Designed Prompt Template.} 
    By giving the concatenated set-of-marks images, the right side (round-1) shows the initial response and the corresponding human edits. Once finalized by humans, these edits will be merged into a single detailed caption grounded with panoptic segmentation masks, as shown in the left side (round-2). 
    }
    \label{fig:template}
\end{figure*}

\section{\textit{COCONut-PanCap} Dataset }
We construct a novel dataset based on COCO images to provide detailed captions at both image and mask levels, using COCONut panoptic masks as a foundation for comprehensive region descriptions. Specifically, we leverage  panoptic masks from COCONut-S~\cite{deng2024coconut} to annotate detailed region captions, incorporating both `thing' and `stuff' masks to cover a wide range of semantic regions. 

\subsection{Dataset Description}
Comprehensively understanding diverse visual elements in complex scenes can benefit multiple tasks including perception, understanding, and generation. In this section, we describe the annotation pipeline for our dataset leveraging the human annotated panoptic masks. We first show the statistical analysis of our final dataset in Tab.~\ref{tab:dataset_comp}. On average, our captions contain 203 words spanning 11 sentences. We follow the same split setting in COCO2017~\cite{lin2014coco} dataset, which includes 118K training images.
To provide a comprehensive evaluation set, we adopt the same 25K images from COCONut-val split (which contains COCO2017-val and another 20K Objects365~\cite{shao2019objects365} validation images).

\subsection{Dataset Construction}
We argue that high-quality descriptions should provide sufficient details of key objects and their attributes, as well as information about secondary objects and background elements. To achieve this, as shown in Fig.~\ref{fig:pipeline}, we use human-annotated panoptic segmentation masks to decide the set of objects to reference in the caption. These masks include both `thing' and `stuff' classes, representing single objects and semantic regions, respectively.
We adopt the panoptic segmentation masks from the COCONut-S~\cite{deng2024coconut} dataset. The masks are overlaid on the images, labeled with class names $c_1, c_2, \dots, c_n \in C$, where $C$ is the set of COCO’s 133 panoptic classes. We then construct a prompt with both the edited image and the original image and a textual question for GPT-4V, as illustrated in Fig.~\ref{fig:template}.
The resulting region captions from GPT-4V are reviewed and corrected by human raters for accuracy and consistency.

\subsection{Dataset Analysis}
\begin{figure}
    \centering
    \includegraphics[width=\linewidth]{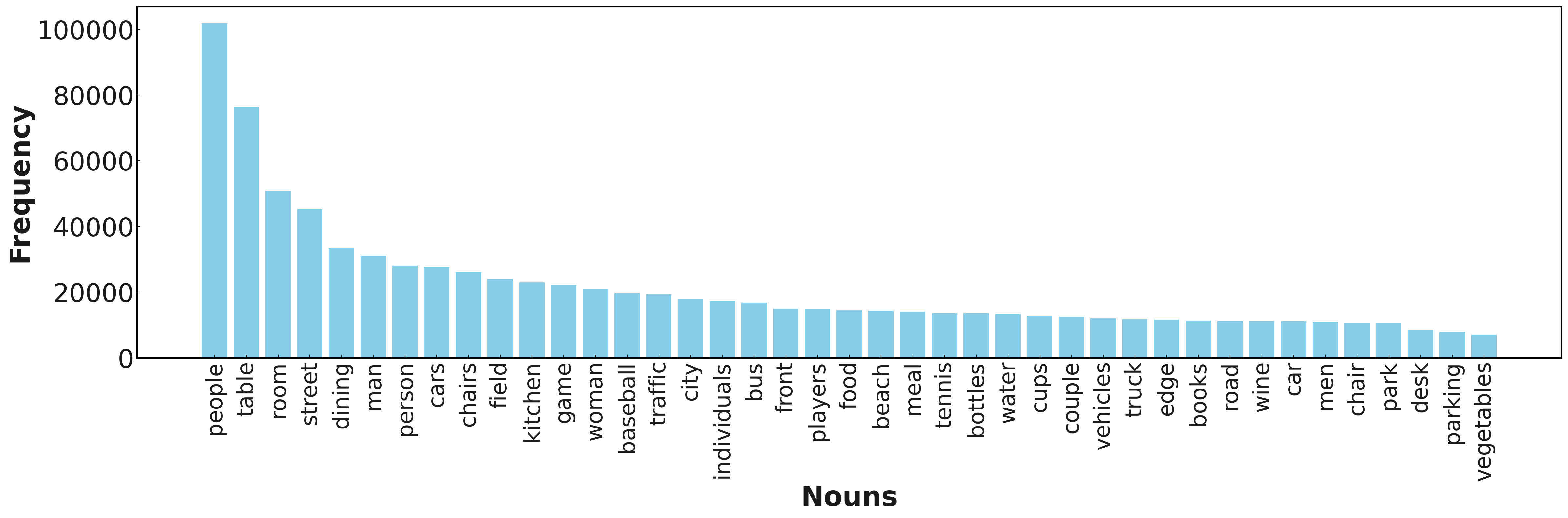}
    \caption{\textbf{Frequency of Extracted Nouns from the COCONut-PanCap Dataset}. The top 10 most frequent nouns are: people, table, room, street, dining, man, person, cars, chairs, and field.}
    \label{fig:freq_nouns}
\end{figure}

\noindent\textbf{Concepts Beyond COCO's 133 Classes.} To clarify the goal of our annotation task, we focus on key visual features such as objects, attributes, spatial relationships, and counting. As shown in Fig.~\ref{fig:freq_nouns}, we utilize the panoptic segmentation mask from COCONut-S, which includes 133 classes in the word vocabulary. Our proposed dataset, however, incorporates additional concepts beyond these 133 classes, such as `vegetable' and `parking'. This demonstrates that our human annotators delivers accurate and diverse descriptions when using the provided label names as a reference.

\begin{figure}[t!]
    \centering
    \includegraphics[width=0.9\linewidth]{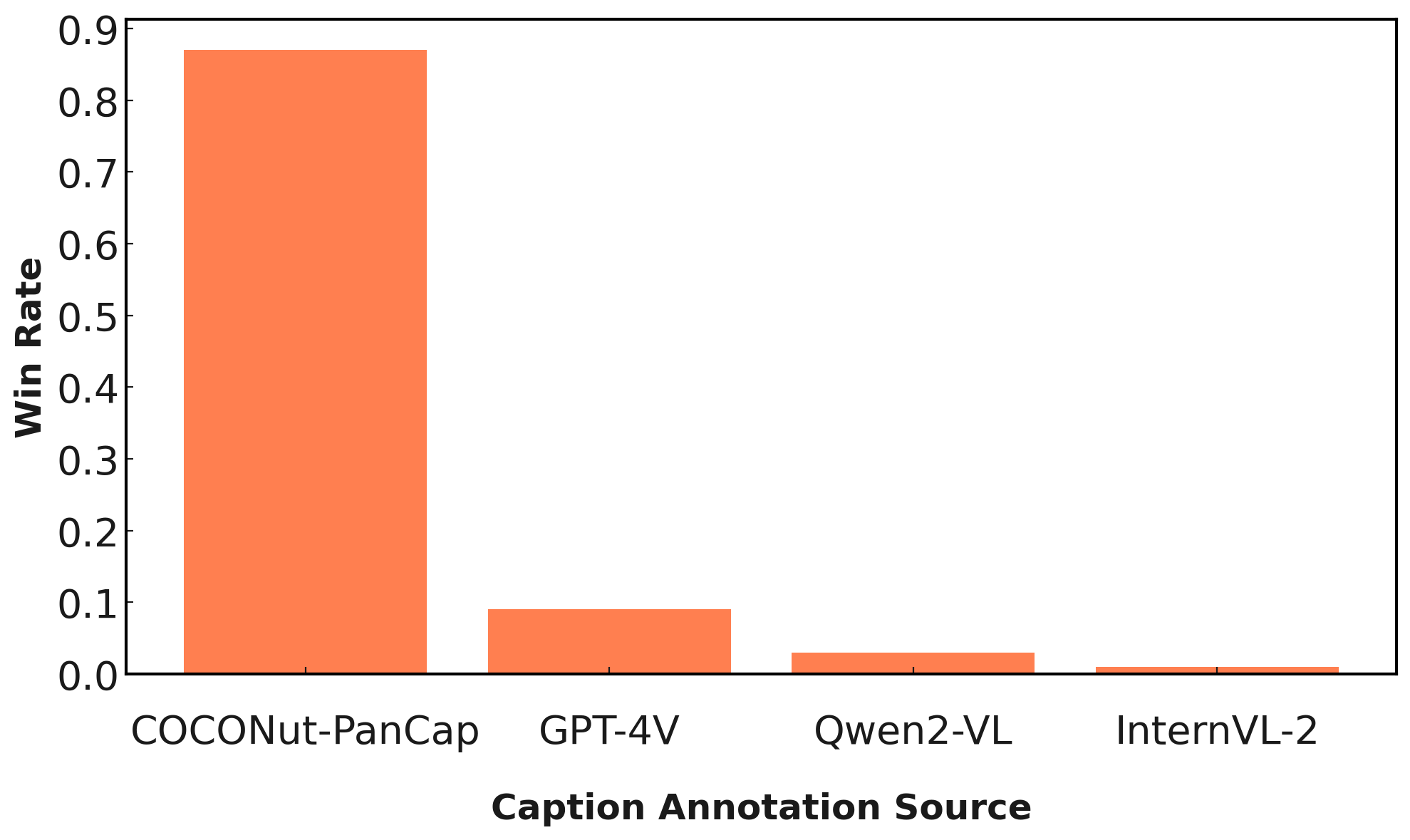}
    \caption{\textbf{Caption Quality via User Study.} The study involved human evaluators assessing a random sample of 1,000 captions, with a strong preference shown for captions from our dataset. }
    \label{fig:user_study}
\end{figure}

\noindent\textbf{User Study for Caption Quality.}
We randomly sample 1,000 images from our COCONut-PanCap training set and asked a human evaluator to perform a single-choice selection task. The question is: \textit{`Please select the best description for the image, considering the correctness of object names, attributes, counting, spatial relationships, and action.}' The compared captions are generated using GPT-4V~\cite{achiam2023gpt4v}, Qwen2-VL~\cite{wang2024qwen2vl}, and InternVL-2~\cite{chen2024internvl2}, resulting in a single-choice four-option question. Fig.~\ref{fig:user_study} illustrates the results, showing that our GPT-assisted human-annotated captions receives the highest ratings. More details can be found in the supplementary.

\section{PGC Baseline: PanCaper }
In this section, we introduce our baseline method for joint panoptic segmentation and grounded captioning (PGC), namely PanCaper. We start with an overview of the pixel grounding task and then present our proposed approach, which incorporates a panoptic segmentation module specifically designed for grounding objects in captions.

\noindent\textbf{Revisiting the Pixel Grounding Task.} Our baseline model builds upon LISA~\cite{lai2024lisa}, a model that combines the language generation capabilities of VLMs with the ability to produce segmentation mask. LISA consists of three main components: a VLM, a vision backbone $V$, and a mask decoder $D$.
With a given text prompt, the VLM (typically LLaVA~\cite{liu2023llava,liu2023improvedllava}) generates an output containing a $\langle \mathrm{SEG} \rangle$ token. For instance, with the input prompt, \textit{`Could you segment the food with high Vitamin C?'} LISA generates the response \textit{`It is $\langle \mathrm{SEG} \rangle$.'} This process extracts the last-layer embedding of the LLM from LLaVA. Then a language-to-prompt (L-P) projection layer ($g$) transforms the last-layer embeddings corresponding to $\langle \mathrm{SEG} \rangle$ tokens ($l_{\mathrm{seg}}$) into the decoder's feature space. Meanwhile, the vision backbone extracts dense visual features from the input image. Finally, both the dense features and the CLIP image embedding from LLaVA are fed into the mask decoder to produce the final segmentation mask.

\noindent\textbf{Prompt Instruction for Grounded Captioning.} We propose a baseline method for the PGC task by modifying LISA to enable grounded captioning with segmentation masks. Since LISA was originally designed for generating segmentation with a single output mask, two main adjustments are necessary: (1) the use of multiple $\langle \mathrm{SEG} \rangle$ tokens, and (2) extracting noun phrases from the caption for grounding.
To facilitate grounded segmentation, we modify the prompt to the VLM as \textit{`Please provide a detailed description of the image and segment each part.'} This prompt triggers the model to generate caption responses with corresponding $\langle \mathrm{SEG_i} \rangle$ tokens, where $i \in [1,N]$ and $N$ is the total number of predicted segmentations. 

Given a predicted caption for the image, aligning each $\langle \mathrm{SEG_i} \rangle$ token requires pairing it with a noun phrase, `$\langle \mathrm{p} \rangle \mathrm{phrase_i} \langle \mathrm{/p} \rangle$,' where $\mathrm{phrase_i}$ is the relevant part in the caption to be grounded.
With these prompt tokens defined, the model uses the vision backbone $V$ and mask decoder $D$ to facilitate fine-grained, pixel-level grounding, with $D$ producing segmentation masks $M$. 

\noindent\textbf{Enable Panoptic Grounding}. To achieve panoptic segmentation from captions, we first classify $\langle \mathrm{SEG} \rangle$ tokens into two types: $\langle \mathrm{SEG_t} \rangle$ for `thing' classes and $\langle \mathrm{SEG_s} \rangle$ for `stuff' classes. These tokens are then processed by our segmentation modules to produce panoptic segmentation masks.
We initialize the vision backbone $V$ with a pretrained kMaX-DeepLab encoder~\cite{yu2022kmaxdeeplab} and fine-tune the decoder $D$ using our COCONut-PanCap dataset. Since kMaX-DeepLab operates as a closed-set segmenter, we align text embeddings of the associated noun phrases with COCO’s 133 panoptic classes. To accomplish this alignment, we use BERT~\cite{koroteev2021bert} to generate the text embeddings and to calculate cosine similarity, selecting the best-matching category.
Panoptic grounding provides mapping between detailed captions and image regions, which  improves interpretability of VLM predictions.

\noindent\textbf{Training Objectives.} Our training objective aims to minimize the following losses:
\begin{equation}
    \mathcal{L} = \lambda_{\text{text}} \mathcal{L}_{\text{text}} + \lambda_{\text{mask}} \mathcal{L}_{\text{mask}},
\end{equation}
where $L_{\text{text}}$ is the auto-regressive cross-entropy loss for text generation, and $L_{\text{mask}}$ is the mask loss~\cite{wang2021max}, encouraging the model to produce high-quality segmentation results.
$\lambda_{\text{text}}$ and $\lambda_{\text{mask}}$ are the respective loss weights. We use the same loss weights as LISA~\cite{lai2024lisa}.

\noindent\textbf{Evaluation Metrics for Caption Quality.} We conduct the analysis with multiple metrics to evaluate the quality and completeness of the generated captions. We introduce a benchmarking suite for the PGC task, with a validation set of 25K images. For the caption quality, we report  the caption metrics including CIDEr~\cite{vedantam2015cider}, METEOR~\cite{banerjee2005meteor}, ROUGE-L~\cite{lin2004rouge}, BLEU\symbol{64}4~\cite{papineni2002bleu} and CAPTURE~\cite{dong2024capture}. For grounded panoptic segmentation, we report PQ scores~\cite{kirillov2019panoptic}.


\begin{table*}[htbp]
\centering
\scalebox{0.95}{
\tablestyle{1.02pt}{1.15}
\begin{tabular}{l|l|c|l|c|c|cccc}

 \textbf{Training recipe} & \textbf{Method}           & \textbf{Pretrain Dataset}    & \textbf{Instruction-tuning dataset}   & \textbf{Mask pooled}       &\textbf{ CAPTURE} & \textbf{CIDEr} &  \textbf{BLEU\symbol{64}4 }    & \textbf{METEOR}   & \textbf{ROUGE-L}   \\

\shline
       \multirow{6}{*}{finetune}                   & LLaVA-NeXT*      & LAION-CC-SBU        & LLaVA 665K            & \xmark                    & 55.4        & 10.8          &   4.2     &   13.2   & 23.1 \\
                          & LLaVA-NeXT       & LAION-CC-SBU        & LLaVA 665K-COCONut-PanCap       & \xmark                    & 58.7        & 11.2          &     4.8   &    16.2  & 24.6 \\
                          & LLaVA-NeXT-pool       & LAION-CC-SBU        & LLaVA 665K-COCONut-PanCap        & \cmark                    & 61.4        & 13.1          & 5.3       &     17.1 & 26.8 \\
                          \cline{2-10}
                          & LLaVA-NeXT-I     & LAION-CC-SBU        & LLaVA 665K-InternVL2-Cap    & \xmark                    & 53.9         & 9.4          &  4.4      & 11.5    &  21.4 \\
                          & LLaVA-NeXT-Q     & LAION-CC-SBU        & LLaVA 665K-Qwen2VL-Cap        & \xmark                    & 55.4         & 8.9         &  4.6      &    12.9  & 22.5 \\
                          & LLaVA-NeXT-G     & LAION-CC-SBU        & LLaVA 665K-GPT4V-Cap        & \xmark                    & 56.2         & 9.6          &  4.7      & 13.3    & 22.8 \\

\end{tabular}%
}
\vspace{-5pt}

\caption{\textbf{Caption Benchmark Results Evaluated on Our COCONut-PanCap Val Set.} Note that the amount of data in the instruction dataset remains the same; only the sources of the detailed captions vary, with a total of 23K images that have detailed captions.}
\label{tab:caption}
\end{table*}

\begin{table*}[htbp]
\centering
\tablestyle{1.2pt}{1.15}
\scalebox{0.95}{
\begin{tabular}{l|cc|c|cccc|ccc}
            &                  &                     & \multicolumn{1}{c}{} & \multicolumn{4}{c}{\textbf{Caption}} & \multicolumn{3}{c}{\textbf{Grounding segmentation} }\\
\textbf{Method}      & \textbf{Pretrain dataset} & \textbf{Instruction dataset} & \textbf{Mask pooled  }     & CAPTURE &  CIDEr& BLEU\symbol{64}4     & METEOR       & PQ     & $\text{PQ}^\text{thing}$  & $\text{PQ}^\text{stuff}$    \\
\shline
LISA+ *       & LAION-CC-SBU     & GranDf      & \xmark                    &   46.2       & 6.6    & 3.8 &   9.8   & 0.43   & 0.41       & 0.45                \\
LISA+      & LAION-CC-SBU     & COCONut-PanCap (ours)      & \xmark                    &     57.9     &  8.1  &  4.9 &   13.8   &  0.50  &   0.49    &  0.44            \\
GLaMM GCG * & LAION-CC-SBU+GranD            & GranDf              & \xmark                    &  43.2       & 6.5       &   3.6   &    10.6  & 0.27   & 0.35       & 0.21               \\
GLaMM GCG   & LAION-CC-SBU+GranD            & COCONut-PanCap (ours)              & \xmark                    & 56.8         & 7.8      &  5.2    &   14.3   & 0.55   & 0.54       & 0.46               \\
PanCaper (ours)      & LAION-CC-SBU     & COCONut-PanCap (ours)     & \xmark                    &   62.6       & 12.0     & 5.8     &    15.4     & 0.56   & 0.55       & 0.66                \\
PanCaper-Pro (ours)     & LAION-CC-SBU     & COCONut-PanCap (ours)     & \cmark                   &   64.3     & 12.5      &   6.4   &    17.9    & 0.61       &      0.58      &     0.68              
\end{tabular}%
}
\vspace{-5pt}

\caption{\textbf{Joint Panoptic Segmentation and Grounded Captioning (PGC) on  COCONut-PanCap Val Set.} * denotes reproduced results.  }
\label{tab:pgc}
\end{table*}

\begin{table}[ht]
\centering
\scalebox{0.9}{
\tablestyle{1.2pt}{1}
\begin{tabular}{l|l|ccc}
\textbf{Training dataset}    & \textbf{Evaluation dataset }   & \textbf{FID$\downarrow$ } &\textbf{ $ \text{FD}_{\text{dinov2}}\downarrow$  }& \textbf{CLIPScore$\uparrow$ }\\
\shline

SD3 PT dataset~\cite{esser2024scaling} &\multirow{4}{*}{DOCCI test set~\cite{Onoe2024docci}}           & 30.2 & 345    & 74.9     \\
COCO-caption~\cite{chen2015coco_caption}  & &27.6 & 321    & 76.8      \\
DOCCI~\cite{Onoe2024docci} & & 22.1 & 300       & 77.8    \\
COCONut-PanCap (ours)&         & 21.4 & 290       & 77.9       \\
\hline
SD3 PT dataset~\cite{esser2024scaling}&  & 31.8 & 300       & 73.8        \\
COCO-caption~\cite{chen2015coco_caption}  & COCONut-PanCap & 28.0    &294  & 74.0      \\
DOCCI~\cite{Onoe2024docci}& val set (ours)       & 24.3 & 267        & 75.1        \\
COCONut-PanCap (ours)&       & 23.1 & 260        & 77.3       \\
\end{tabular}
}
\vspace{-5pt}

\caption{\textbf{Benchmark Results on Text Conditioned Image Generation.}  Stable-Diffusion-3 (SD3) medium is finetuned with COCO-Caption (short), DOCCI and our COCONut-Panoptic and evaluated on DOCCI test set~\cite{Onoe2024docci} and our COCONut-PanCap val set. `SD3 PT dataset' denotes the pretraining dataset of SD3, and thus the rows correspond to zero-shot evaluation of SD3.} 
\label{tab:t2i}
\end{table}
\begin{table}[ht!]
\centering
\scalebox{0.9}{
\tablestyle{1.02pt}{1.15}
\begin{tabular}{l|cccc}
        & \textbf{w/o FT} & 
        \textbf{COCO-caption~\cite{chen2015coco_caption} }& \textbf{DOCCI~\cite{Onoe2024docci}} & \textbf{COCONut-PanCap }\\
\shline
color attribution & 0.37         & 0.34            & 0.38     & 0.40                     \\
colors            &  0.73        & 0.70           & 0.74     & 0.75                            \\
position          &  0.33         & 0.30            & 0.36     & 0.36                            \\
counting         & 0.65        & 0.64            & 0.65     & 0.70                            \\
single object     &  0.96      & 0.94            & 0.95     & 0.96                             \\
two objects       & 0.80        & 0.78            & 0.81    &  0.89                             \\
\hline
overall score     &  0.64       & 0.62          &  0.65    & 0.68                            
\end{tabular}
}
\vspace{-5pt}

\caption{
\textbf{Effects of Fine-tuning the SD3-medium (T2I model) with Different Datasets on GenEval~\cite{ghosh2023geneval}.} w/o FT denotes the model is not finetuned with any datasets (\ie, zero-shot testing).
}
\label{tab:geneval}
\end{table}

\begin{table*}[htbp]
\centering
\tablestyle{1.5pt}{1.05}
\begin{tabular}{l|c|l|ccccccc}

\textbf{Method   }              & \textbf{LLM }      & \textbf{Instruction-tuning Dataset}       & \textbf{MM-Vet} & \textbf{Seed-IMG} & \textbf{MMBench-en} & \textbf{TextVQA} & \textbf{POPE} & \textbf{MME}  \\
\shline
LLaVA-NeXT * & Llama3-8B & orginal LLaVA 665K~\cite{liu2024llavanext}               & 43.5   & 70.1     & 71.4        & 68.9    & 85.4 & 1523 \\
LLaVA-NeXT-20K                & Llama3-8B & LLaVA 665K-COCONut-PanCap-20K               & 44.1   & 72.5     & 73.6        & 69.8    & 86.1 & 1552 \\
LLaVA-NeXT-50K                & Llama3-8B & LLaVA 665K-COCONut-PanCap-50K                     & 44.6   & 73.1     & 74.2        & 70.0    & 87.1 & 1600 \\
LLaVA-NeXT-Full               & Llama3-8B & LLaVA 665K-COCONut-PanCap-118K                  & 45.5   & 74.3     & 75.1        & 70.7    & 87.9 & 1612 \\
\hline

LLaVA-1.5  & Vicuna-7B & LLaVA 665K-ShareGPT4V-100K & 37.8 & 67.4 &70.5 & 64.6 & 84.7& 1519 \\
 
LLaVA-1.5  & Vicuna-7B & LLaVA 665K-COCONut-PanCap-20K  & 38.5 & 67.7 &70.9 &64.5 & 84.9 & 1521 \\
\end{tabular}
\caption{\textbf{Benchmark Results and Ablation Study on VQA.}  By adding extra detailed caption data for instruction tuning, the models show increased improvement. * denotes reproduced results. Using only \textbf{20K human labeled data} can still achieve \textbf{comparable performance} to 100K synthetic data.
}
\label{tab:ablate_vqa}
\end{table*}

\begin{table*}[ht]
\centering
\scalebox{0.95}{
\begin{tabular}{l|ccc|ccc|cc}
\tablestyle{1.02pt}{1.15}
\multirow{2}{*}{\textbf{Method}} & \multicolumn{3}{c}{\textbf{refCOCO}  }                                                   & \multicolumn{3}{c}{\textbf{refCOCO+}  }                                                  & \multicolumn{2}{c}{\textbf{refCOCOg}   }                    \\
                        & \multicolumn{1}{c}{val} & \multicolumn{1}{c}{testA} & \multicolumn{1}{c}{testB} & \multicolumn{1}{c}{val} & \multicolumn{1}{c}{testA} & \multicolumn{1}{c}{testB} & \multicolumn{1}{c}{val} & \multicolumn{1}{c}{test} \\
                        \shline

\textcolor{lightgray}{GLaMM*~\cite{hanoona2023GLaMM}} & \textcolor{lightgray}{77.5} & \textcolor{lightgray}{79.2}    & \textcolor{lightgray}{74.9}    & \textcolor{lightgray}{71.3} & \textcolor{lightgray}{74.7}    & \textcolor{lightgray}{61.5}    & \textcolor{lightgray}{71.3} & \textcolor{lightgray}{71.9}     \\
PixelLM~\cite{ren2024pixellm}                 & 73.0                      & 76.5                      & 68.2                      & 66.3                    & 71.7                      & 58.3                      & 69.3                    & 70.5                     \\

LISA-7B~\cite{lai2024lisa}                 & 74.1                    & 76.5                      & 71.1                      & 62.4                    & 67.4                      & 56.5                      & 66.4                    & 68.5                     \\
PanCaper$^{+}$                    & 74.5                    & 76.7                      & 69.9                      & 69.9                    & 73.4                      & 59.5                      & 69.8                    & 70.6                     \\

PanCaper$^{+}$ +  COCONut-PanCap    & 76.2                    & 77.1                      & 72.3                      & 70.5                    & 73.9                      & 60.1                      & 72.1                    & 71.6                    
\end{tabular}
}
\vspace{-5pt} 
\caption{\textbf{Benchmark Results on Referring Segmentation.} * denotes reproduced results. It is noted that GLaMM uses extra data from the GranD dataset for pretraining. $^{+}$ denotes our PanCaper model is adapted for referring segmentation task.}
\label{tab:refcocco}
\end{table*}

\section{Experimental Results}
We assess the effectiveness of human-annotated caption data by performing three primary tasks utilizing our dataset in the fine-tuning/instruction tuning stage: detailed captioning, panoptic grounded captioning (PGC), and text-to-image generation. Additionally, we demonstrate the transferability of the knowledge learned from our dataset through two downstream tasks: VQA and referring segmentation.

\noindent\textbf{Detailed Captioning.} We conduct instruction tuning with LLaVA-NeXT framework~\cite{liu2024llavanext} for this task. We replace the caption data (23k) from the original LLaVA instruction-tuning set with detailed captions from our dataset, keeping the same amount of instruction data size. We follow the same training setup used for LLaVA-NeXT with Llama3-8B~\cite{dubey2024llama3}. Treating it as a QA task, we use the prompt, \textit{`Could you please describe the image in detail?'} and collect the corresponding response as the caption for the image. We evaluate caption quality using CIDEr~\cite{vedantam2015cider}, METEOR~\cite{banerjee2005meteor}, BLEU@4~\cite{papineni2002bleu}, ROUGE-L~\cite{lin2004rouge} and CAPTURE~\cite{dong2024capture} metrics. We also extend the model by adding the mask-pooled features from the panoptic segmentation masks as additional signals to the LLaVA model and name it LLaVA-NeXT-pool.  During training, we use the ground truth mask to extract the features while during inference we use the mask proposals from the pretrained kMaX-DeepLab~\cite{yu2022kmaxdeeplab}. Besides, we also experiment with synthetic captions directly generated using InternVL-2~\cite{chen2024internvl2}, Qwen2-VL~\cite{wang2024qwen2vl} and GPT-4V~\cite{achiam2023gpt4v}. We follow the same data preparation settings as our dataset to build these instruction datasets for these 23K images with different sources of synthetic detailed captions, namely LLaVA 665K-InternVL2-Cap , LLaVA 665K-Qwen2VL-Cap, and LLaVA 665K-GPT4V-Cap. These datasets are used to produce models LLaVA-NeXT-I, LLaVA-NeXT-Q, and LLaVA-NeXT-G respectively.
More details can be found in the supplementary. The results are presented in Tab.~\ref{tab:caption}. LLaVA-NeXT models show improved performance when fine-tuned on the custom instruction-tuning dataset. Among these, LLaVA-NeXT-pool achieves the highest scores in all metrics, with CAPTURE of 61.4, CIDEr of 13.1, BLEU@4 of 5.3, and METEOR of 17.1, significantly higher than the original model variant LLaVA-NeXT, indicating the benefit of added region features for additional visual cues.  Models trained on synthetic captions (LLaVA-NeXT-I, LLaVA-NeXT-Q, and LLaVA-NeXT-G) generally show lower scores, showing advantage of our human-annotated caption.

\noindent\textbf{PGC: Stronger Detail Reasoning Performance.} We implement our proposed PanCaper based on LISA which uses pre-trained LLaVA-NeXT with a LLM of Llama3-8B, with LoRA~\cite{hu2021lora} adopted. The vision encoder uses a fixed CLIP-ViT-L/14-336 model, modified with linearly interpolated position embeddings to process 448 resolution images. The trainable components of our model include the mask decoder of kMaX-DeepLab, and the tunable parts in LLaVA same as in LISA. To enhance model performance in visual understanding, we initialize our PanCaper using pretrained LLaVA-NeXT models from the detailed captioning task. We also experiment with a model variant that uses mask pooled features similar to LLaVA-NeXT-pool, and name it PanCaper-Pro.

For comparison, we select 3 related methods LISA, PixelLM~\cite{ren2024pixellm} and GLaMM~\cite{hanoona2023GLaMM} for evaluation. It is noteworthy that LISA is not able to perform multi-mask prediction. We therefore adapt LISA~\cite{lai2024lisa} for the multi-mask generation with grounded segmentation, namely LISA+. The implementation details can be found in the supplementary. Tab.~\ref{tab:pgc} shows the quantitative results. Our proposed PanCaper-Pro achieves the highest scores across all captioning metrics (CIDEr: 12.5, CAPTURE: 64.3, BLEU@4: 6.4, METEOR: 17.9), outperforming all other models. Both PanCaper models show significant improvements over other models in all captioning metrics, highlighting the effectiveness of the COCONut-PanCap dataset for detailed caption generation. On grounding segmentation, PanCaper-Pro again leads, with a PQ score of 0.61, $\text{PQ}^\text{thing}$ of 0.58, and $\text{PQ}^\text{stuff}$ of 0.68, reflecting its robustness on both ``thing'' and ``stuff'' classes. Notably, enabling mask pooling in our proposed PanCaper-Pro further enhances segmentation metrics. The baseline models (LISA+ and GLaMM with GranD) achieve much lower PQ scores, due to incomplete segmentation annotations in the GranD dataset.

\noindent\textbf{Text-to-Image Generation.} We adopt the Stable Diffusion 3 (SD3) medium model\footnote{\scriptsize https://huggingface.co/docs/diffusers/stable\_diffusion/stable\_diffusion\_3} for text to image generation with LoRA finetuning. We adopt the default training settings but only with different text-image datasets for training. We evaluate with two types of training images from COCO~\cite{lin2014coco} and DOCCI~\cite{Onoe2024docci} datasets. In details, for the COCO images, we explore the short COCO-caption and detailed captions from our dataset. For DOCCI images, we directly use the captions from their dataset. Tab.~\ref{tab:t2i} shows the quantitative results. Traning on COCONut-PanCap achieves the best performance across all metrics when evaluated on DOCCI-test, with the lowest FID (21.4), lowest $\text{FD}_{\text{dinov2}}$ (290), and the highest CLIPScore (77.9), indicating superior generation quality and high image-text relevance. 
When evaluated on COCONut-PanCap-val set, training on COCONut-PanCap again shows the best results with the lowest FID (23.1), $\text{FD}_{\text{dinov2}}$ (267), and a high CLIPScore of 77.3.

Tab.~\ref{tab:geneval} shows the results on GenEval benchmark~\cite{ghosh2023geneval}. Finetuning SD3-medium with COCONut-PanCap consistently scores the highest in most categories, particularly those requiring image details like color attribution, object positioning, and handling multiple objects. Our proposed dataset enables more accurate image generation that requires understanding of relationships, multiple objects and counting, tasks that other datasets struggle with.

\noindent\textbf{VQA.} To evaluate the effectiveness of the proposed COCONut-PanCap dataset, we utilize these captions during the instruction-tuning stage and follow the setup of LLaVA-NeXT~\cite{liu2024llavanext} across various visual question answering (VQA) and multi-modality understanding benchmarks. We evaluate on MM-Vet~\cite{yu2024mm-vet}, SEED-IMG~\cite{li2023seed}, MMBench-en~\cite{liu2023mmbench}, MME~\cite{fu2023mme}, POPE~\cite{li2023pope}, and TextVQA~\cite{singh2019textvqa}, covering a broad range of evaluation dimensions. We experiment with different amount of our COCONut-PanCap caption data injected into the instruction tuning stage by replacing the original COCO captioning data with our dataset. As shown in Tab.~\ref{tab:ablate_vqa}, the baseline model LLaVA-NeXT (using its original recaptioned COCO) achieves relatively lower performance across all metrics, with scores such as 43.5 on MM-Vet, 70.1 on Seed-IMG, and 68.9 on TextVQA. 
Building on LLaVA-NeXT baseline, we progressively incorporated varying amounts of COCONut-PanCap data (20K, 50K, and 118K (full), as indicated by postfixes in the baseline names) during instruction-tuning. Consistent improvements are observed across all evaluated benchmarks as more of our data is integrated.

\noindent\textbf{Referring Segmentation.} In this task, the model processes an image and a textual referring expression to output a segmentation mask corresponding to the expression. The prompt used is, \textit{`Please segment the $\langle \mathrm{referring\_text} \rangle$ in the image.'} The target model response is \textit{`Sure, it is  $\langle \mathrm{SEG} \rangle$.'}, where the  $\langle \mathrm{SEG} \rangle$ token is decoded to obtain the mask. We follow the setup in LISA~\cite{lai2024lisa}, using multiple segmentation datasets to jointly train the models. 
Tab.~\ref{tab:refcocco} shows the quantitative results. Our model achieves superior performance, particularly when additionally trained with the COCONut-PanCap dataset (last row), outperforming all models except GLaMM~\cite{hanoona2023GLaMM}. This improvement underscores our model's efficacy in handling complex referring expressions, likely due to the additional data that enhances model generalization and accuracy. It is worth noting that GLaMM performs competitively with our method, though the comparison is uneven given their additional use of the SA-1B dataset~\cite{kirillov2023segmentanything}.

\begin{figure}[htbp]
    \centering
    \includegraphics[width=0.9\linewidth]{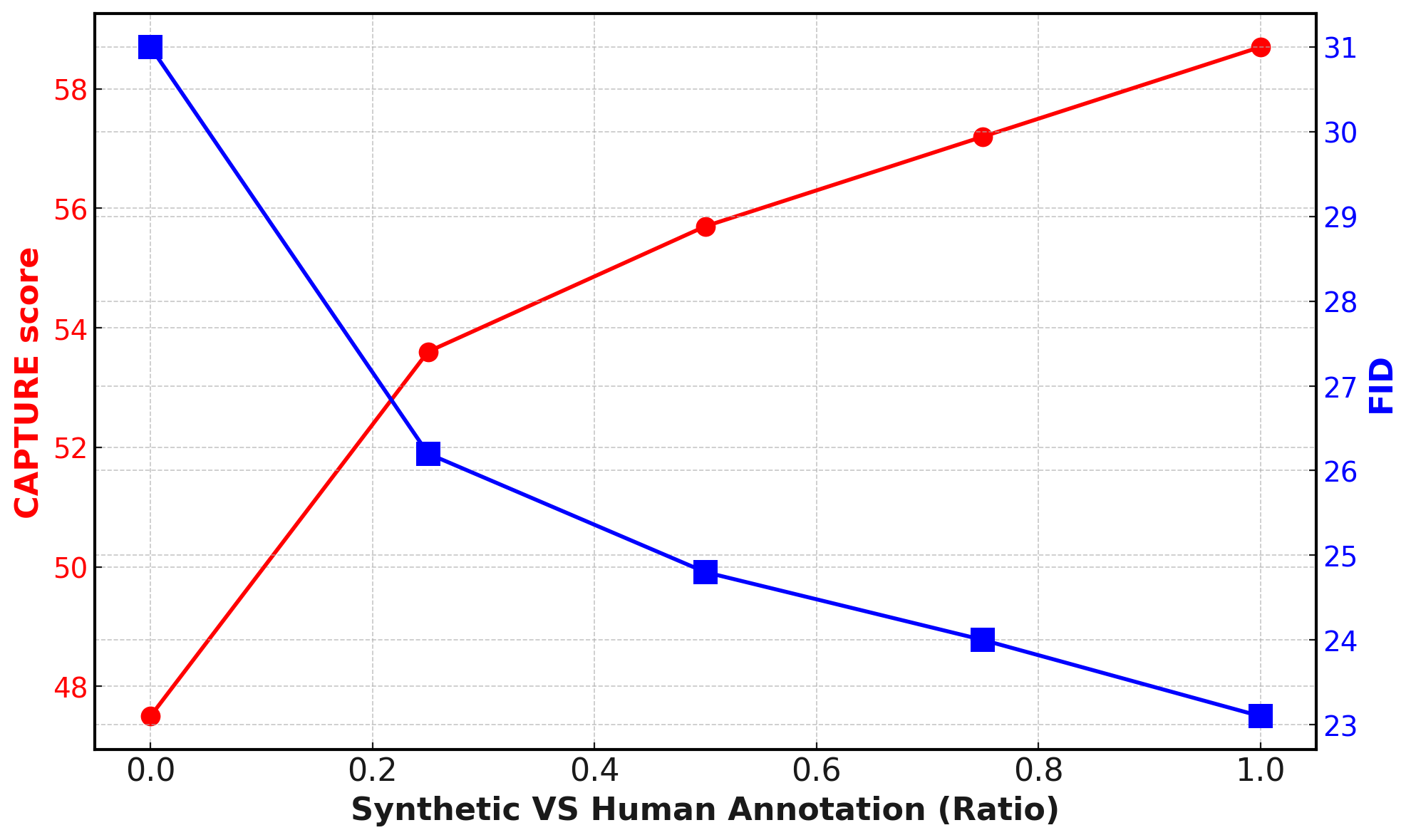}
    \vspace{-5pt}
    \caption{
    \textbf{Varying Synthetic and Human-Annotation Ratios.}
    CAPTURE is used to evaluate the performance of LLaVA-NeXT on detailed captioning, while FID assesses the performance of SD3-medium on text-conditioned image generation.
    }
        \vspace{-10pt}
    \label{fig:fid_capture}
\end{figure}

\noindent\textbf{Synthetic \vs Human Annotated Data.} Generating synthetic data for captioning has been popular for recent tasks in either training vision encoders~\cite{radford2021clip} or text-to-image generation~\cite{li2024recaption}. We investigate the effect of varying the mix ratio of synthetic captions generated by GPT-4V and our human-annotated data for fine-tuning (where 0 indicates fully synthetic data), using the COCONut-PanCap dataset for training and the COCONut-PanCap-val set for evaluation. We adopt LLaVA-NeXT for the captioning task and SD3-medium for the image generation task. As shown in Fig.~\ref{fig:fid_capture}, adding 25\% human-annotated data yields significant performance improvements in both captioning and generation, with a reduced FID of 26 from 31 (lower is better) and an increased CAPTURE score of 53.6 from 47.5 (higher is better). Consistent improvements are observed as more human-annotated data is incorporated.

\section{Conclusion and Discussion}

In this work, we proposed a novel dataset designed to support detailed captioning and grounded segmentation tasks built on COCO images. We demonstrated that our dataset can enhance model performance during instruction tuning and fine-tuning stages across various multi-modal understanding and generation tasks, such as captioning, VQA, grounded segmentation, and text-to-image generation. We hope that COCONut-PanCap, with its detailed captions grounded with dense panoptic masks, will foster future advancements in multi-modal learning research.

\noindent\textbf{Limitations.}
High-quality human-labeled data offers significant benefits for instruction tuning in multi-modal tasks, but scaling such datasets is challenging. To address this, we introduce  COCONut-PanCap as a starting point for large-scale human-annotated data exploration. 
Recognizing the relatively smaller dataset size compared to other large dataset, future work may involve using this dataset to train seed models to generate more high-quality synthetic data.

{
    \small
    \bibliographystyle{ieeenat_fullname}
    \bibliography{main}
}

\clearpage

\appendix

\noindent \textbf{The appendix is organized as follows.}
\begin{itemize}
    \item In Sec.~\ref{sec:experimental_details}, we show implementation details for Detailed Captioning (Sec.~\ref{subsec:detailed_captioning}), Panoptic segmentation and Grounded (Sec.~\ref{subsec:pgc}), and VQA (Sec.~\ref{subsec:vqa}).
    \item In Sec.~\ref{sec:qualitative_results}, we show more visualization examples of our proposed COCONut-PanCap dataset (Sec.~\ref{subsec:data}), and analysis of the tier cases in our dataset annotation user study (Sec.~\ref{subsec:tier}).
\end{itemize}

\section{Experimental Details}
\label{sec:experimental_details}
In this section, we provide more experimental details for detailed captioning (Sec.~\ref{subsec:detailed_captioning}), PGC (Sec.~\ref{subsec:pgc}), and VQA (Sec.~\ref{subsec:vqa}).

\subsection{Detailed Captioning}
\label{subsec:detailed_captioning}

\noindent\textbf{Detailed Captioning Instruction Dataset Construction.}
The key step in conducting the experiment is constructing the dataset. The original LLaVA-665K dataset consists of LLaVA-158K combined with other VQA datasets. Within LLaVA-158K, a subset of detailed captions corresponds to 23K COCO images. To create our-LLaVA-665K (referred to as LLaVA 665K-COCONut-PanCap in the table), we replace the detailed caption annotations for these 23K COCO images with our annotations. Importantly, the total amount of training data remains unchanged (only the captions for these 23K images are updated), ensuring a fair comparison of the impact of data quality on model performance.

\noindent\textbf{Synthetic Annotation for Detailed Caption.} To build the synthetic dataset with state-of-the-art VLM, we use three models, including open-sourced InterVL-2, Qwen2-VL and close-sourced GPT-4V to generate the detailed captions for COCO 118K train set images. We use the same text prompts that is used in LLaVA~\cite{liu2023llava} for prompting the model to create the detailed captions.

\begin{figure}[t!]
    \centering
    \begin{subfigure}[b]{0.49\textwidth}
        \centering
        \includegraphics[width=\textwidth]{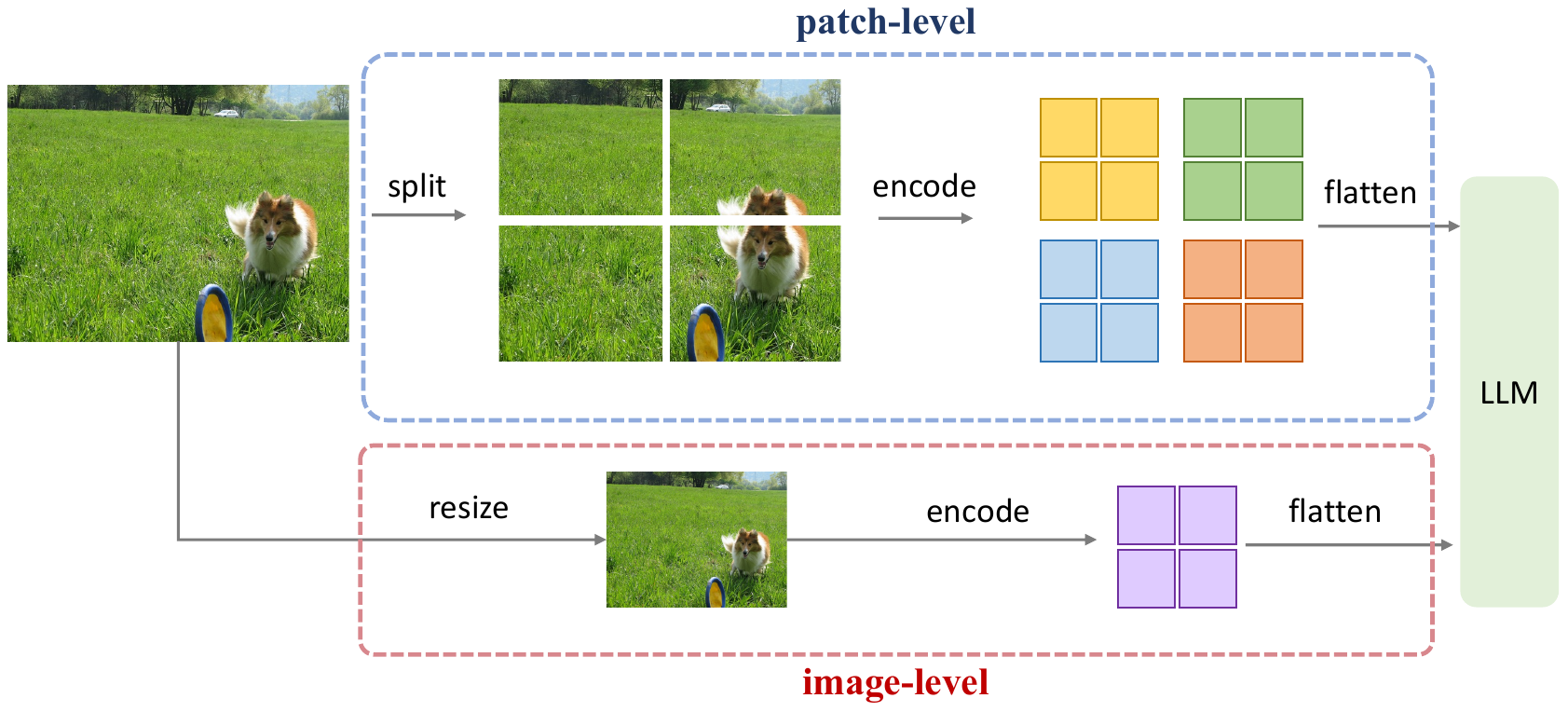}
        \caption{LLaVA-NeXt-AnyRes}
        \label{fig:llava_anyres}
    \end{subfigure}
    \hfill
    \begin{subfigure}[b]{0.49\textwidth}
        \centering
        \includegraphics[width=\textwidth]{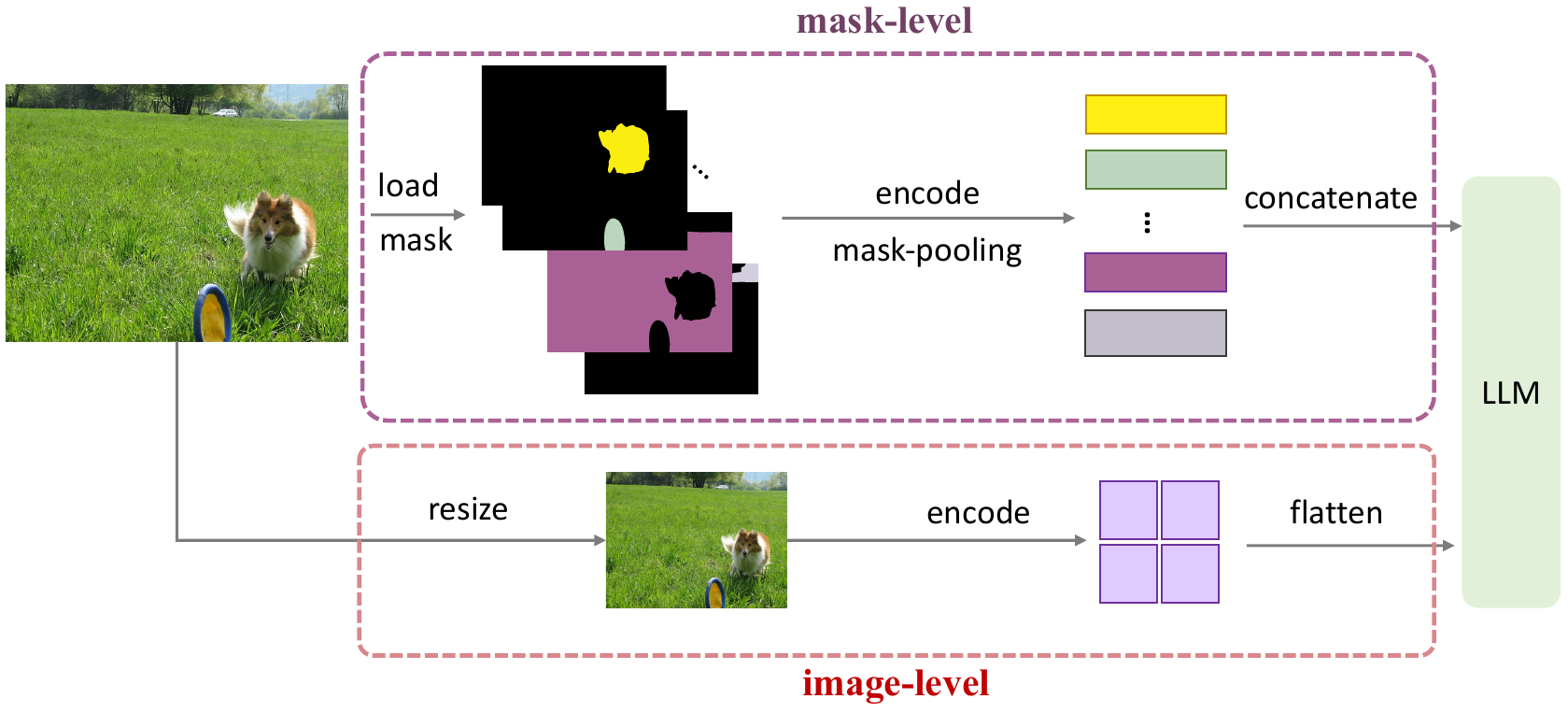}
        \caption{our LLaVA-NeXt-pool}
        \label{fig:llava_pool}
    \end{subfigure}
    \caption{\textbf{Comparison of LLaVA-NeXt and our proposed LLaVA-NeXt-pool.}
    }
    \label{fig:llava_arch}
\end{figure}

\noindent\textbf{LLaVA-NeXT-pool implementation details.} Fig.~\ref{fig:llava_arch} shows the comparison of the original LLaVA-NeXT and our proposed LLaVA-NeXT-pool. As shown in Fig.~\ref{fig:llava_anyres}, in order to preserve the details for the high-resolution images and representations, the original design employs a grid configuration which can also balance the performance efficiency with operational costs. Then both the patch-level and image-level features are later concatenated and sent to the LLM. Directly splitting the image into patches could cause prolems, for example, in the figure, the upper part of the dog's head is partitioned into different patches which may result in incomplete feature extraction for single object. To overcome this drawback, we propose LLaVA-NeXT-pool to extract the dense feature and preserve the object details by utilizing the panoptic segmentation masks in our COCONut-PanCap dataset. Fig.~\ref{fig:llava_pool} shows the details. Compared to the original design, LLaVA-NeXT-pool could effectively extract the features for the dog in our example. Our design enables more complete region-level feature extraction and is potential in understanding the details better.

\begin{figure*}[ht!]
    \centering
    \includegraphics[width=\linewidth]{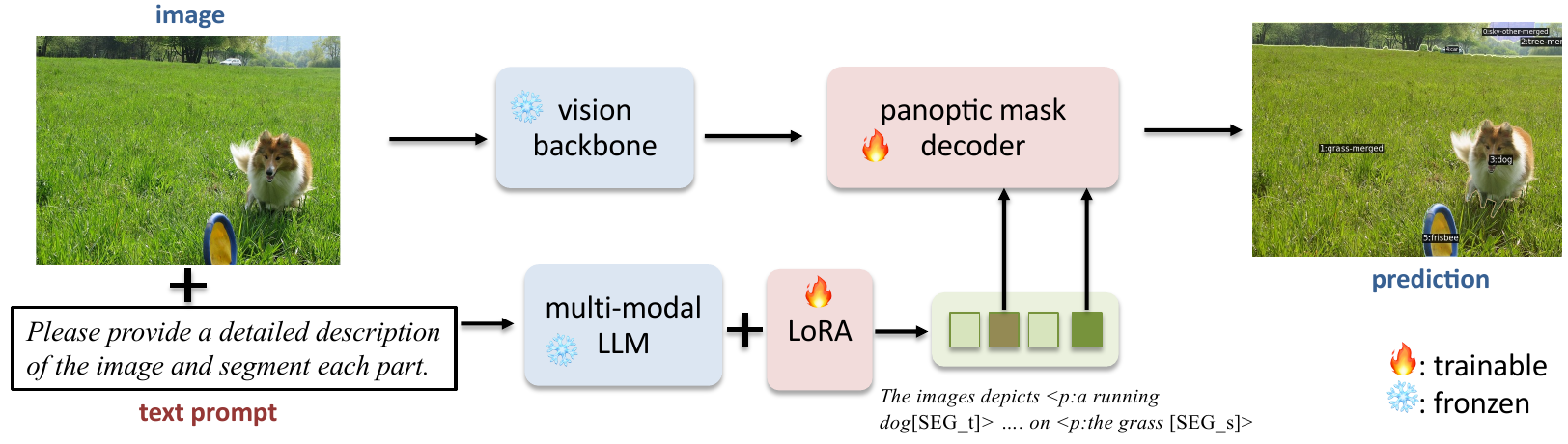}
    \caption{\textbf{Architecture of PanCaper.} We utilize a pretrained vision encoder from kMaX-DeepLab~\cite{yu2022kmaxdeeplab} as our vision backbone, which effectively extracts dense features essential for panoptic segmentation.}
    \label{fig:pancaper_arch}
\end{figure*}

\subsection{PGC}
\label{subsec:pgc}
We provide more implementation details for the proposed task: \textbf{P}anoptic segmentation and \textbf{G}rounded \textbf{C}aptioning (PGC).

\noindent\textbf{PanCaper Implementation Details.} We introduce the PanCaper architecture details in this section. Following the architecture in LISA~\cite{lai2024lisa}, there are three components including the vision backbone, mask decoder and multi-modal LLM. Fig.~\ref{fig:pancaper_arch} shows the architecture details for PanCaper. We made modification on the vision backbone, and mask decoder part in terms of model architecture. To preserve the learned knowledge of the pre-trained multimodal LLM (\ie, LLaVA-NeXT in our experiments), we leverage LoRA~\cite{hu2021lora} to perform efficient fine-tuning, and completely freeze the vision backbone. The mask decoder is fully fine-tuned. Additionally, the LLM token embeddings (embed tokens), the LLM head (lm head), and the projection layer are also trainable.
The weights of the text generation loss $\lambda_{\text{text}}$  and the mask
loss $\lambda_{\text{mask}}$ are set to 1.0 and 1.0, respectively. For the PQ-style mask loss, we follow the same settings in kMaX-DeepLab~\cite{yu2022kmaxdeeplab}, where it consists of mask-level cross entropy loss, dice loss and pixel loss.

\noindent\textbf{Adapting Baseline Methods for PGC Task.}
We adopt the same text prompt template to enable the model to perform PGC tasks. For LISA$+$, we follow the same design in GLaMM~\cite{hanoona2023GLaMM} to design the multi entity mask output by utilizing the the GranDf dataset. As the intruction dataset of GranDf is constructed similarly grounding the phrase in the image-level caption, it will output multiple $\langle \mathrm{SEG} \rangle$ tokens. The reasoning results of the number of $\langle \mathrm{SEG} \rangle$ tokens decide the number of output entity mask which are often binary masks. As a result, the model can generate a detailed caption along
with interleaved segmentation masks, employing the format ``$\langle \mathrm{p} \rangle$A man$\langle \mathrm{/p} \rangle$$\langle \mathrm{SEG} \rangle$ ... next to $\langle \mathrm{p} \rangle$a tree$\langle \mathrm{/p} \rangle$$\langle \mathrm{SEG} \rangle$''. And thus the format of instruction dataset is significat in task design. Therefore, we formulate our dataset as ``$\langle \mathrm{p} \rangle$A man$\langle \mathrm{/p} \rangle$$\langle \mathrm{SEG_t} \rangle$ ... next to $\langle \mathrm{p} \rangle$a tree$\langle \mathrm{/p} \rangle$$\langle \mathrm{SEG_s} \rangle$'', where $\langle \mathrm{SEG_t} \rangle$ represents the seg token for instance masks of thing and $\langle \mathrm{SEG_s} \rangle$ represents for semantic masks of stuff respectively in panoptic setting. Similarly, utilizing the PanCap dataset and special token design, GLaMM~\cite{hanoona2023GLaMM} is able to generate the entity masks with the tag of `thing' and `stuff'.

\noindent\textbf{Training Data Formulation.} We adopt the same training data from LISA~\cite{lai2024lisa} which comprises mainly three parts, all of which are derived from widely-used public datasets. These include 1) Semantic Segmentation datasets including  ADE20K~\cite{zhou2017ade20k}, COCO-Stuff~\cite{caesar2018coco_stuff}, and LVIS-PACO~\cite{ramanathan2023paco} part datasets with the generated QA data, 2) Vanilla Referring Segmentation Datasets: refCOCO, refCOCO+, refCLEF~\cite{kazemzadeh2014refcoco} and refCOCOg~\cite{mao2016refcocog} datasets, 3) ReasonSeg dataset~\cite{lai2024lisa}, and 4) Visual Question Answering Dataset: LLaVA-v1.5-mix665k~\cite{liu2023improvedllava}. To enable the multi-mask generation for grounded caption, there are two options for instruction datasets, GranDf and our COCONut-PanCap where GranDf consists of entity masks while COCONut-PanCap consists of panoptic masks.

\subsection{VQA}
\label{subsec:vqa}
We provide more implementation details for the VQA experiments. We follow the same setting in LLaVA-NeXT to create the experimental results for VQA tasks. We focus on the instruction tuning stage by adopting the pretrained weights from the stage-1 across the trainings for all the model variants mentioned in Tab.~7 in the paper. The dataset we used is exactly the same as in LLaVA 665K~\cite{liu2023improvedllava} which includes the earlier version of instruction data proposed in LLaVA 158K~\cite{liu2023llava}, ShareGPT~\cite{sharegpt}, VQAv2~\cite{mao2016vqav2}, GQA~\cite{hudson2019gqa}, openknowledge VQA (OKVQA~\cite{marino2019okvqa}, A-OKVQA~\cite{schwenk2022aokvqa}), OCR (OCRVQA~\cite{mishra2019ocrvqa}, TextCaps~\cite{sidorov2020textcaps}), region-level VQA datasets (Visual Genome~\cite{krishna2017visualgenome}, RefCOCO~\cite{kazemzadeh2014refcoco}). Among these data, LLaVA 158K comprises 77K complex reasoning, 58K conversation and 23K detailed captions. To build the dataset variants shown in Tab.~7, we simply remove the subset of detailed\_caption\_23k, and subsequently add 20K, 50K and 118K COCONut-PanCap dataset to build LLaVA 665K-COCONut-PanCap-20K, LLaVA 665K-COCONut-PanCap-50K and LLaVA 665K-COCONut-PanCap-118K. By these steps, we add more detailed caption data to construct the instruction tuning dataset. This results in the total amount of training data of 662K for LLaVA 665K-COCONut-PanCap-20K, 692K for LLaVA 665K-COCONut-PanCap-50K and 760K for LLaVA 665K-COCONut-PanCap-118K. And thus the size of LLaVA 665K-COCONut-PanCap-20K is slightly smaller than the original LLaVA 665K dataset, but the model trained on it yields better performance. For the evaluation settings, we follow the exact settings in LLaVA-NeXT~\cite{liu2024llavanext} using lmms\_eval\footnote{https://github.com/EvolvingLMMs-Lab/lmms-eval}.

\section{More Qualitative Results}
\label{sec:qualitative_results}
In this section, we present additional qualitative results of COCONut-PanCap annotations (Sec.~\ref{subsec:data}) and a detailed analysis of tier cases from the user study (Sec.~\ref{subsec:tier}).

\subsection{Data Examples}
\label{subsec:data}
\begin{figure*}
    \centering
    \includegraphics[width=\linewidth]{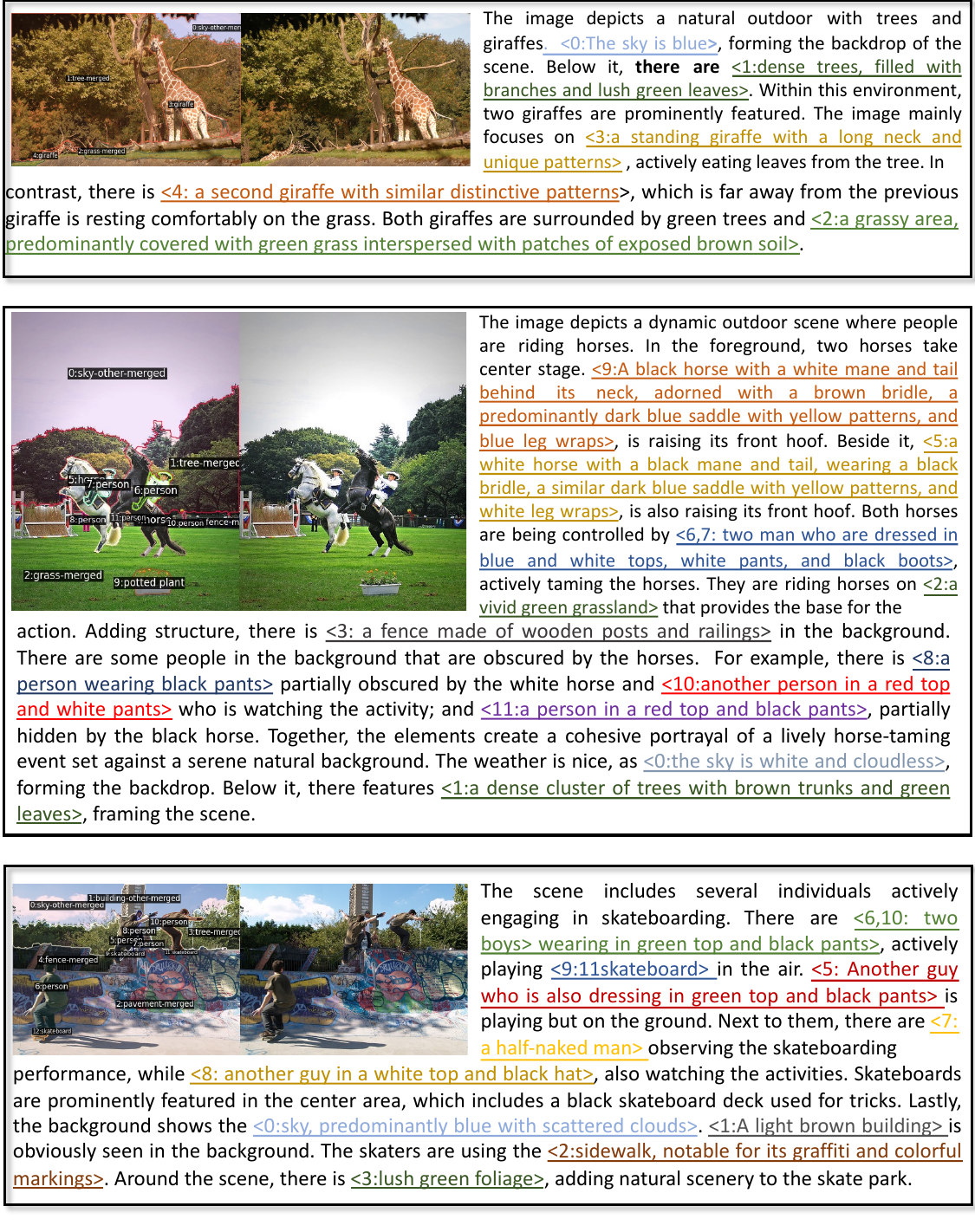}
    \caption{
    \textbf{Visualization of the Panoptic Grounded Caption.}
    Our annotated captions ground the panoptic segmentation masks.
    }
        \vspace{-20pt}

    \label{fig:vis_examples_1}
\end{figure*}

\begin{figure*}
    \centering
    \includegraphics[width=0.95\linewidth]{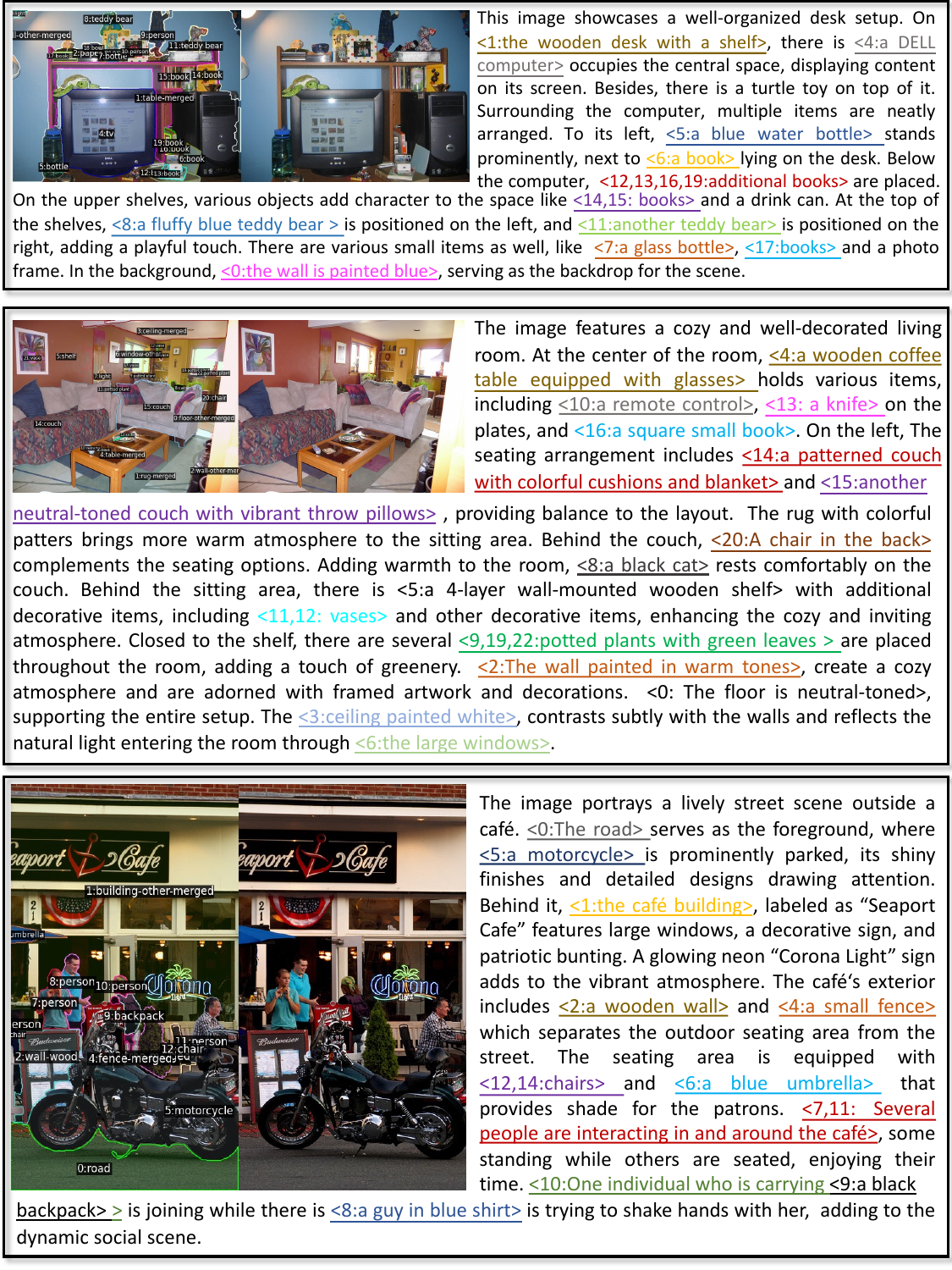}
    \caption{
    \textbf{Visualization of the Panoptic Grounded Caption.}
    Our annotated captions ground the panoptic segmentation masks.
    }
        \vspace{-25pt}

    \label{fig:vis_examples_2}
\end{figure*}

We show more visualization of our proposed COCONut-PanCap dataset in Fig.~\ref{fig:vis_examples_1} and Fig.~\ref{fig:vis_examples_2}.

\subsection{PanCaper and GPT-4V Tier Showcases}
\label{subsec:tier}
\begin{figure*}
    \centering
    \includegraphics[width=0.95\linewidth]{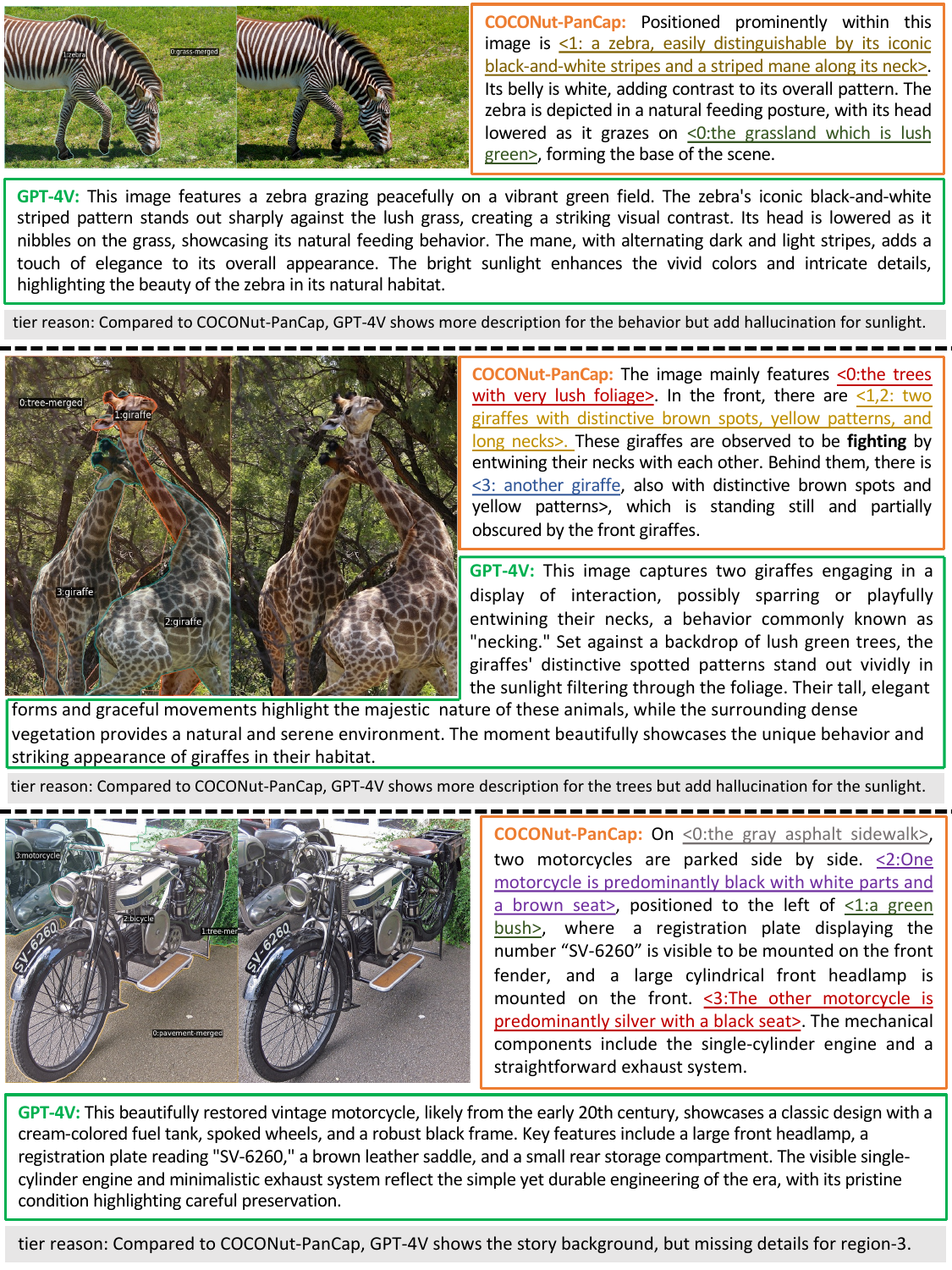}
        \vspace{-10pt}

    \caption{
    \textbf{Tier Examples for the User Study.}
    Our COCONut-PanCap annotations are tied with GPT-4V annotations for some simple cases.
    }
    \label{fig:tier_examples_1}
\end{figure*}

\begin{figure*}
    \centering
    \includegraphics[width=0.95\linewidth]{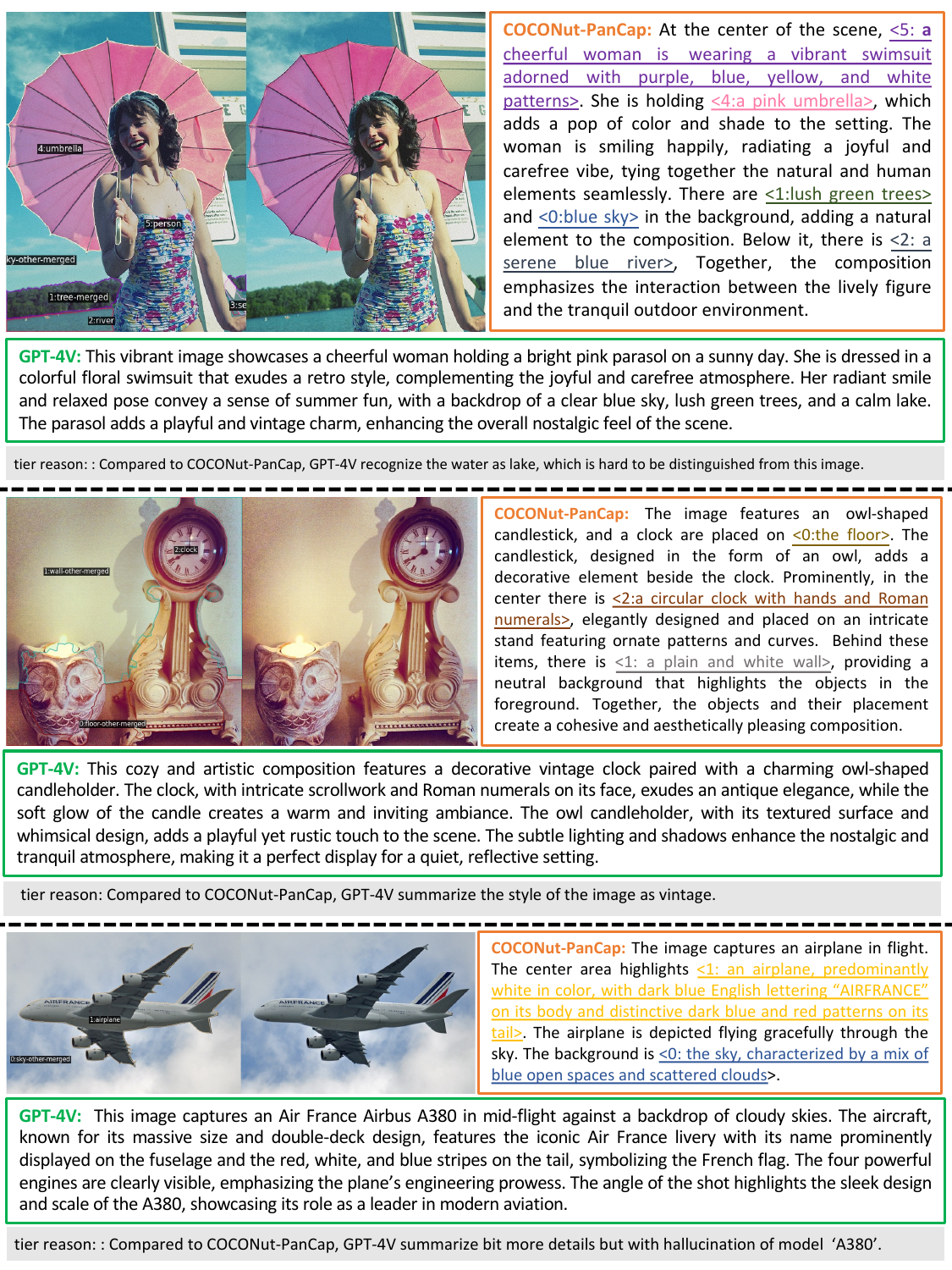}
    \vspace{-10pt}
    \caption{
    \textbf{Tier Examples for the User Study.}
    Our COCONut-PanCap annotations are tied with GPT-4V annotations for some simple cases.
    }
    \label{fig:tier_examples_2}
\end{figure*}

\begin{figure*}
    \centering
    \includegraphics[width=0.95\linewidth]{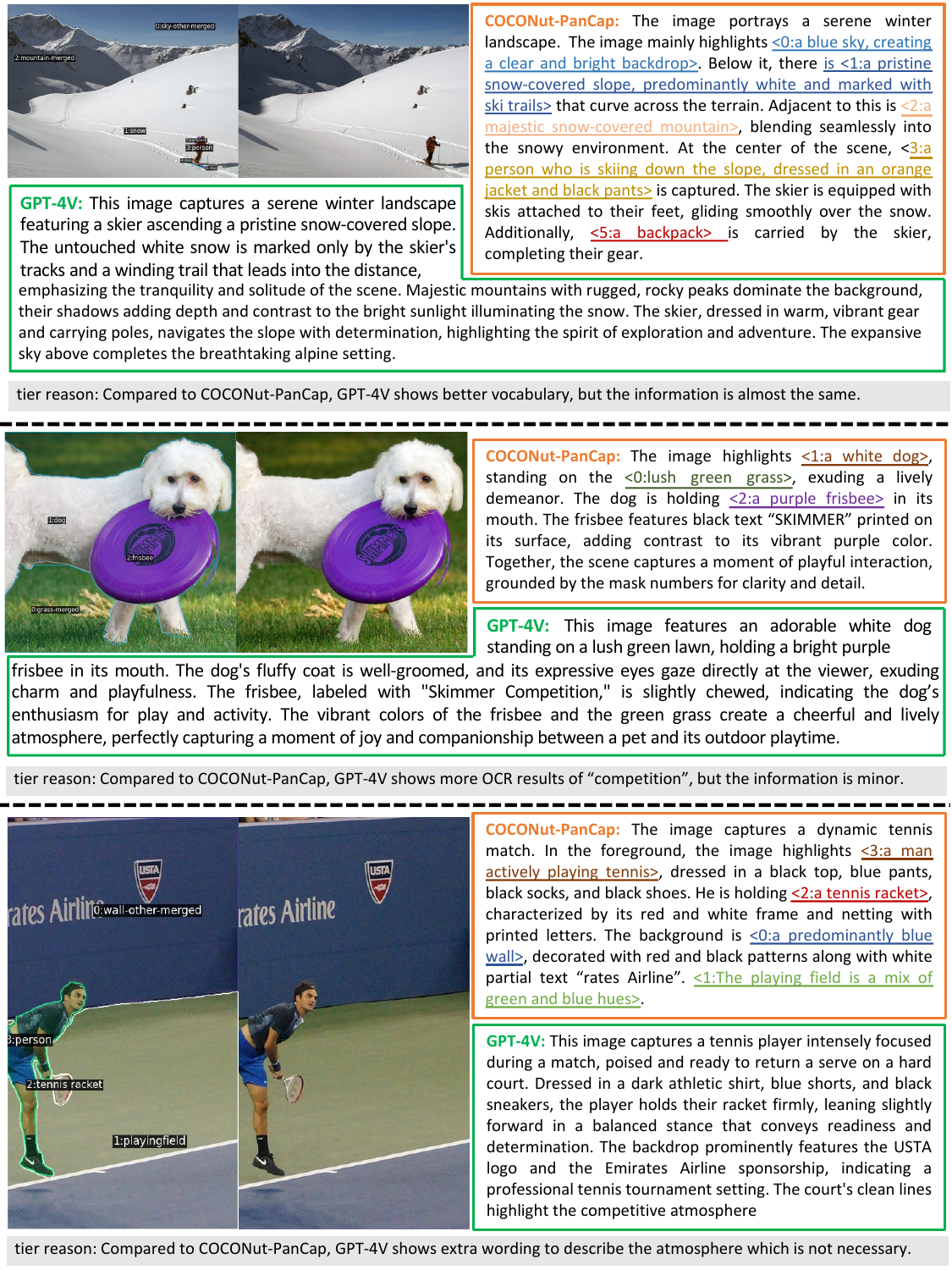}
        \vspace{-10pt}

    \caption{
    \textbf{Tier Examples for the User Study.}
    Our COCONut-PanCap annotations are tied with GPT-4V annotations for some simple cases.
    }
    \label{fig:tier_examples_3}
\end{figure*}

In the user study involving 1,000 samples, captions generated by GPT-4V were preferred in 87 cases. Among these, actually, 46 were tier cases where human raters considered both GPT-4V and COCONut-PanCap captions equally good. Fig.~\ref{fig:tier_examples_1}, Fig.~\ref{fig:tier_examples_2} and Fig.~\ref{fig:tier_examples_3} illustrate qualitative examples, highlighting the reasons for the tier classification and instances where GPT-4V was chosen.

\end{document}